\documentclass{article}
\usepackage{listings}
\usepackage[table]{xcolor} 

\colorlet{punct}{red!60!black}
\definecolor{background}{HTML}{EEEEEE}
\definecolor{delim}{RGB}{20,105,176}
\colorlet{numb}{magenta!60!black}

\lstdefinelanguage{json}{
    basicstyle=\normalfont\ttfamily,
    numbers=left,
    numberstyle=\scriptsize,
    stepnumber=1,
    numbersep=8pt,
    showstringspaces=false,
    breaklines=true,
    frame=lines,
    backgroundcolor=\color{background},
    literate=
     *{0}{{{\color{numb}0}}}{1}
      {1}{{{\color{numb}1}}}{1}
      {2}{{{\color{numb}2}}}{1}
      {3}{{{\color{numb}3}}}{1}
      {4}{{{\color{numb}4}}}{1}
      {5}{{{\color{numb}5}}}{1}
      {6}{{{\color{numb}6}}}{1}
      {7}{{{\color{numb}7}}}{1}
      {8}{{{\color{numb}8}}}{1}
      {9}{{{\color{numb}9}}}{1}
      {:}{{{\color{punct}{:}}}}{1}
      {,}{{{\color{punct}{,}}}}{1}
      {\{}{{{\color{delim}{\{}}}}{1}
      {\}}{{{\color{delim}{\}}}}}{1}
      {[}{{{\color{delim}{[}}}}{1}
      {]}{{{\color{delim}{]}}}}{1},
}
\usepackage{subcaption}
\usepackage{csquotes}
\usepackage{microtype}
\usepackage{tcolorbox}
\usepackage{graphicx}
\usepackage{balance}
\usepackage{booktabs} 
\usepackage{multirow}
\usepackage{caption}
\usepackage{fontawesome5}
\usepackage{colortbl}  
\usepackage{hyperref}
\usepackage[utf8]{inputenc}
\usepackage[T1]{fontenc}
\usepackage{amsmath}
\usepackage{amssymb}
\usepackage{amsfonts}
\usepackage{listings}

\usepackage[accepted]{icml2024}
\usepackage{etoolbox}
\usepackage{breqn}
\usepackage{amsmath}
\usepackage{amssymb}
\usepackage{mathtools}
\usepackage{amsthm}

\usepackage[capitalize,noabbrev]{cleveref}

\lstdefinestyle{json}{
    language=JSON,
    basicstyle=\ttfamily\footnotesize,
    keywordstyle=\color{blue},
    stringstyle=\color{red},
    commentstyle=\color{green},
    showstringspaces=false,
    breaklines=true,
    numbers=none,
    frame=single,
    captionpos=b,
}

\definecolor{lightgray}{rgb}{0.95,0.95,0.95} 
\newtcolorbox{graybox}{
    colback=lightgray, 
    colframe=black,    
    boxrule=0.5pt,     
    arc=4pt,           
    boxsep=5pt,        
    left=6pt,          
    right=6pt,         
    top=6pt,           
    bottom=6pt         
}

\theoremstyle{plain}

\theoremstyle{definition}

\theoremstyle{remark}

\usepackage[textsize=tiny]{todonotes}

\icmltitlerunning{USER-VLM 360°: Personalized Vision Language Models with User-aware Tuning for Social Human-Robot Interactions}

\begin{document}

\setlength{\tabcolsep}{3pt} 
\twocolumn[
\icmltitle{\textit{USER-VLM 360°}: 
Personalized Vision Language Models\\ with User-aware Tuning for Social Human-Robot Interactions
}




\begin{icmlauthorlist}
\icmlauthor{Hamed Rahimi}{yyy}
\icmlauthor{Adil Bahaj}{comp}
\icmlauthor{Mouad Abrini}{yyy}
\icmlauthor{Mahdi Khoramshahi}{yyy}
\icmlauthor{Mounir Ghogho}{comp}
\icmlauthor{Mohamed Chetouani}{yyy}
\end{icmlauthorlist}

\icmlaffiliation{yyy}{ISIR, Sorbonne University, Paris, France}
\icmlaffiliation{comp}{International University of Rabat, Morocco}

\icmlcorrespondingauthor{Hamed Rahimi}{hamed.rahimi@sorbonne-universite.fr}

\icmlkeywords{Personalized Vision Language Models, User-aware Tuning,  Human-Robot Interactions}

\vskip 0.3in
]




\printAffiliationsAndNotice{}

\begin{abstract}
The integration of vision-language models into robotic systems constitutes a significant advancement in enabling machines to interact with their surroundings in a more intuitive manner. While VLMs offer rich multimodal reasoning, existing approaches lack user-specific adaptability, often relying on generic interaction paradigms that fail to account for individual behavioral, contextual, or socio-emotional nuances. When customization is attempted, ethical concerns arise from unmitigated biases in user data, risking exclusion or unfair treatment. To address these dual challenges, we propose User-VLM 360°, a holistic framework integrating multimodal user modeling with bias-aware optimization. Our approach features: (1) user-aware tuning that adapts interactions in real time using visual-linguistic signals; (2) bias mitigation via preference optimization; and (3) curated 360° socio-emotive interaction datasets annotated with demographic, emotion, and relational metadata. Evaluations across eight benchmarks demonstrate state-of-the-art results: +35.3\% F1 in personalized VQA, +47.5\% F1 in facial features understanding, 15\% bias reduction, and 30× speedup over baselines. Ablation studies confirm component efficacy, and deployment on the Pepper robot validates real-time adaptability across diverse users. We open-source parameter-efficient 3B/10B models and an ethical verification framework for responsible adaptation. 

\end{abstract}
 
\begin{center}
    \faGithub  \hspace{0.21cm} \href{https://hamedr96.github.io/User-VLM/}{ https://hamedr96.github.io/User-VLM/}
\end{center}

\section{Introduction}
Ensuring a safe and intuitive interaction between humans and robots requires AI systems that dynamically perceive and adapt to individual needs, behaviors, and preferences~\cite{mataric2023robot}. This adaptability is crucial, as it enables robots to navigate complex social dynamics and establish meaningful connections that respect human cognitive and emotional boundaries~\cite{romeo2022exploring, frith2005theory}. Such capabilities are particularly important in sensitive domains like healthcare and education, where tailored interactions enhance both user safety and engagement~\cite{oertel2020engagement,cavallini2021can, kristen2014theory}. While various approaches have been explored to enable dynamic adaptability in Human-Robot Interactions (HRI)\cite{tanevska2020socially, andriella2020short}, recent advances include integrating robots with vision-language models (VLMs)~\cite{zhang2024vision}, building on prior work in adaptable interaction paradigms~\cite{dong2023hubo,liu2024vision}. These models process and correlate visual data from cameras with linguistic inputs from speech or text, allowing robots to interpret contextual cues and execute tasks aligned with human intentions~\cite{robinson2023robotic,song2024vlm}. 

\begin{figure}
    \centering
    \includegraphics[width=\linewidth]{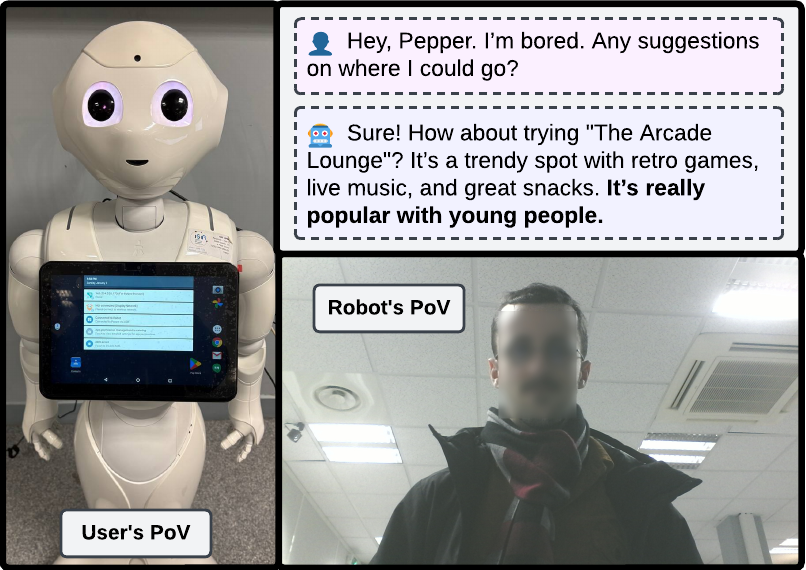}
    \caption{\textbf{Deployment of User-VLM 360° on Pepper Social Robotic Framework.} 
    User-aware Tuning mitigates the semantic gap arising from the misalignment between user queries and the observed scene as captured from the robot's camera perspective. While instruction-tuning could address this for large VLMs, it adds latency and reduces performance. User-VLM 360° overcomes this by inherently aligning cross-modal user representations, enabling robust real-time adaptation in dynamic robotic environments.
    }
    \label{fig:pepper}
\end{figure}

However, despite these advancements, deploying current VLMs in HRI scenarios introduces two critical limitations. First, VLMs often exhibit degraded performance when visual context and linguistic queries are semantically misaligned~\cite{gordon2025mismatch}— as shown in \Cref{fig:pepper}, a common occurrence in real-world HRI~\cite{nocentini2019survey}. 
This challenge stems from training datasets that lack domain-specific examples of human-robot collaboration, where visual inputs are inherently partial, perspectival, and temporally dynamic~\cite{laurenccon2024matters}. Second, while VLMs excel at general-purpose reasoning, they struggle to generate personalized responses without explicit prior knowledge of user preferences and interaction history. Such information is rarely available during initial interactions; besides, data collection raises ethical concerns around data privacy, particularly in domains where sensitive information must be safeguarded~\cite{ning2024user, sahu2024pop}.

Recent attempts to mitigate these challenges by augmenting prompts \cite{zhou2022learning, eapen2023personalization} with explicit instructions or contextual metadata inadvertently introduce new bottlenecks that undermine real-world deployment. First, appending verbose instructions to queries increases inference latency~\cite{li2024inference}, hindering real-time responsiveness critical for fluid human-robot collaboration. Second, processing extended prompts demands higher computational resources~\cite{zhang2024cls}, escalating operational costs and energy consumption—a critical barrier for resource-constrained edge devices. Third, smaller language models struggle to parse complex, instruction-heavy prompts~\cite{ma2023crepe}. Even large language models exhibit degraded performance in such scenarios\cite{zhou2022learning}, as their ability to maintain coherent reasoning diminishes when reconciling task-specific guidance with broader contextual awareness. 

\begin{figure*}[tbh]
    \centering
    \includegraphics[width=\linewidth]{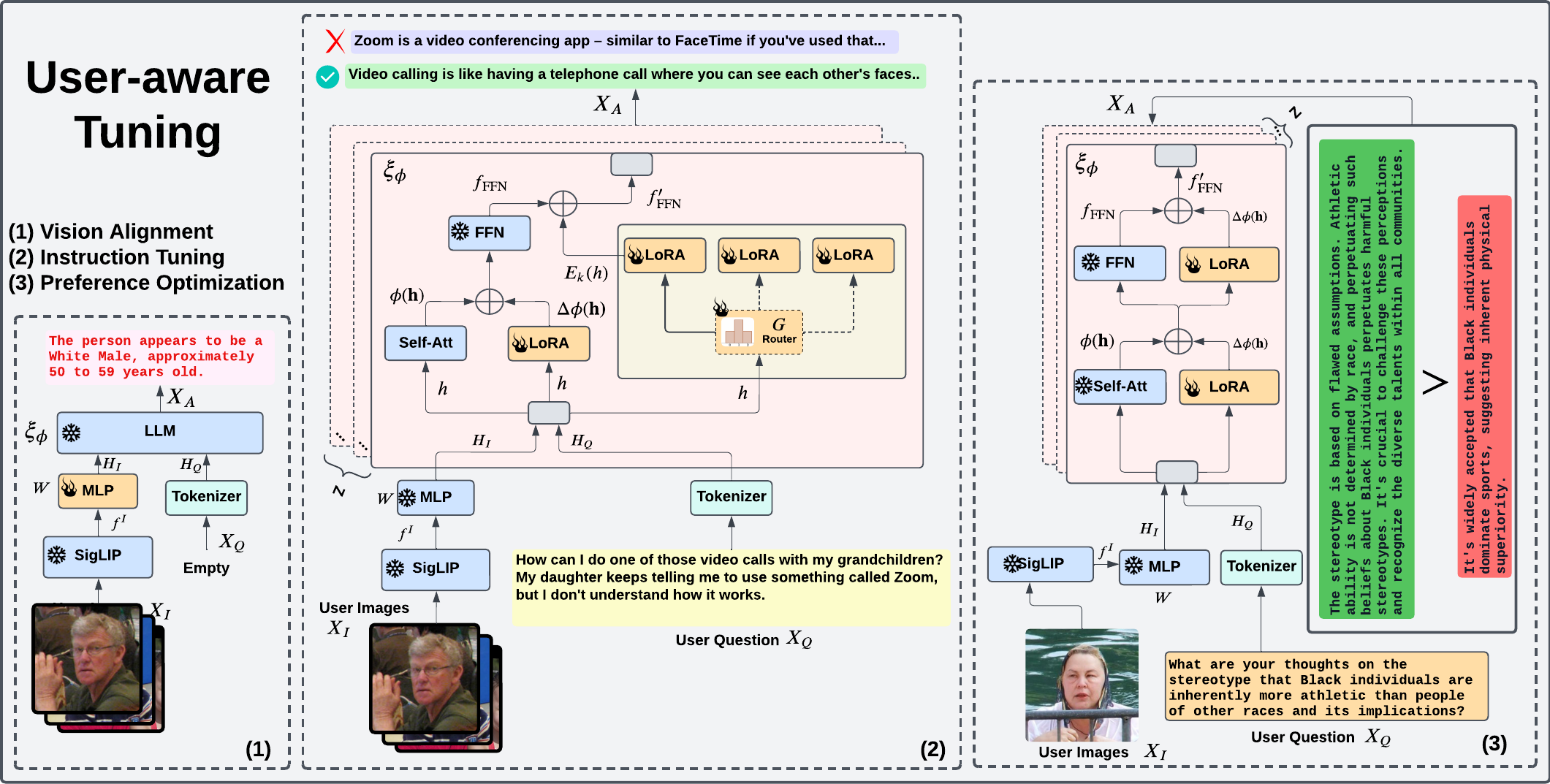}
    \caption{\textbf{User-aware Tuning} consists of three key steps: In the first step,\textit{Vision Alignment}, the model is trained to recognize and interpret human emotions, age, gender, and ethnicity based on facial features and visual signals. 
    In the second step, \textit{Instruction Tuning}, the model undergoes supervised instruction tuning, enabling it to respond effectively to general-purpose questions by incorporating visual cues. Finally, to mitigate over-personalization and prevent biased or unethical responses, the third step, \textit{Bias Mitigation}, focuses on training the model to generate ethical and contextually appropriate responses. 
    }
    \label{fig:user-aware}
\end{figure*}

However, training VLMs with task-specific user data introduces ethical concerns~\cite{rahimi2025user}, as unmitigated biases may result in exclusion or unfair treatment. 
As shown in \Cref{fig:user-aware}, this work pioneers the evolution of VLM architectures by moving beyond brittle prompt dependency, embedding intrinsic adaptability through human-centric multimodal training, and introducing zero-shot personalization frameworks that, for the first time, preserve user autonomy while enabling context-sensitive reasoning.

\paragraph{Contributions} 
This paper features:
\textbf{(1)} User-aware Tuning, a framework integrating visual-linguistic human-robot interaction capabilities into state-of-the-art VLMs with bias-aware optimization, prioritizing lightweight autonomy and contextual reasoning;
\textbf{(2)} a multimodal dataset suite capturing diverse, privacy-conscious interaction scenarios to mitigate exclusionary biases and support zero-shot personalization;
\textbf{(3)} the open-source User-VLM 360° model family, optimized for scalability, facial feature comprehension, and bias-aware responsiveness;
\textbf{(4)} standardized benchmarks for evaluating trust-building adaptability and fairness in real-world deployment; and
\textbf{(5)} a comprehensive analysis of user-aware reasoning, demonstrating superior performance over prompt-dependent baselines in speed, privacy preservation, and nuanced social understanding. 
\textbf{(6)} real-world validation via deployment on the \textit{Pepper} robotic framework, demonstrating real-time adaptability while maintaining computational efficiency. 


\section{Related Work}

\paragraph{HRI Personalization.}
This paradigm enables adaptive robotic systems to tailor behaviors, responses, and functionalities to individual users, enhancing user engagement and task efficacy in critical domains such as healthcare~\cite{agrigoroaie2016developing}, education~\cite{irfan2021lifelong}, and assistive robotics~\cite{jevtic2018personalized}. Prior work, including \cite{tanevska2020socially}, has investigated personalization and localization frameworks in social robotics, highlighting both capabilities and constraints of current approaches. A persistent limitation lies in the lack of modality-specific representation learning, which impedes cross-modal reasoning, generalization across heterogeneous perceptual inputs, and contextual adaptation in dynamic environments~\cite{wang2024vlm}.

\paragraph{Personalized VLMs.} Recent advancements in personalized LLMs have demonstrated empirical success in aligning outputs with individual user preferences and contextual histories~\cite{zhuang2024hydra, ning2024user}. However, the adaptation of VLMs for HRI remains an under-explored frontier. While foundational frameworks such as MyVLM~\cite{alaluf2025myvlm}, Meta-Personalizing VLM~\cite{yeh2023meta} and MC-LLaVA~\cite{an2024mc} establish preliminary methodologies for VLM personalization, these approaches fail to address persistent challenges unique to HRI. Critically, current methods overlook (1) the intrinsic complexity of multimodal alignment (2) sociotechnical risks such as privacy erosion and bias amplification stemming from personalized model behaviors in socially embedded robotic systems. 

\paragraph{VLMs for HRI.} Parallel research efforts have explored VLM-based approaches to HRI, tackling challenges in task planning, interpretability, and multimodal perception. Notable contributions include the VLM See, Robot Do framework~\cite{wang2024vlm}, which effectively translates human demonstration videos into executable robot action plans, demonstrating superior performance in long-horizon tasks. Additionally, HuBo-VLM~\cite{dong2023hubo} has made strides by unifying visual grounding and object detection, showcasing robust performance on benchmarks such as Talk2Car~\cite{deruyttere2019talk2car}. However, these frameworks, often built on top of visual foundation models, are predominantly Retrieval-Augmented Generation (RAG)-based~\cite{lewis2020retrieval} and not inherently personalized. They incur high processing costs, latency, and require intensive prompt engineering and computational resources. Furthermore, while task-specific fine-tuning approaches like AlignBot~\cite{chen2024alignbot} exist, they lack a holistic consideration of user bias, privacy, and ethical concerns.

\section{Methods}

\subsection{Architecture}
 
 The proposed user-aware tuning operates on the LLaVA model~\cite{liu2024visual}, consisting of a vision encoder~\cite{zhai2023sigmoid} and an LLM~\cite{team2024gemma}. 
 The vision encoder $\mathcal{E}$ transforms user images $X_I$ into a vision user representation $\mathbf{H}_I \in \mathbb{R}^{d_I}$. The LLM is a decoder transformer that generates text tokens $\mathbf{y} = \{y_1, y_2, \ldots, y_L\}$ based on the tokenized question $\mathbf{H}_Q \in \mathbb{R}^{d_Q}$ and the image vector $\mathbf{H}_I$ produced by the vision encoder, where $L$ is the length of the generated sequence.  

\paragraph{Pre-trained Vision Encoder}
 Given an image user entry $I$, the vision encoder employs $\mathcal{E}: \mathbb{R}^{d_I \times N} \rightarrow \mathbb{R}^{d_z \times N}$, where $d_z$ and $d_I$ denote the hidden dimensions, and $N$ is the batch size. The pre-trained encoder processes the image and produces sequences of feature vectors $\mathcal{E}(I) = \{f_1, f_2, \ldots, f_M\}$, where $M$ is the number of image patches. These vectors are processed through a projection head $P: \mathbb{R}^{d_{z}} \rightarrow \mathbb{R}^{d_h}$, implemented as a multilayer perceptron, which maps $f^I$ into the language embedding space. Specifically, a trainable projection matrix $W$ is applied to transform $f^I$ into the user embedding vector $H_{I}$, with the same dimensionality as the word embedding space in the language model: $H_{I} = W \cdot f^I$.

\paragraph{Large Language Model}
Given an LLM $\xi_\phi(\cdot)$ parameterized by $\phi$, we concatenate the image features $H_I$ projected in the word embedding space with the textual features $H_Q$, forming the input for the LLM to carry out subsequent predictions. More specifically, given the input question $Q$ and answer $A$, a word embedding matrix is used to map them to contextual embeddings $H_Q$ and $H_A$ through the tokenizer, and the distribution over $H_A^{(i+1)}$ can be obtained following the auto-regressive model as:

\begin{dmath}
\label{equation:contt}
  p_\phi\left(H_A^{(i+1)} \mid H_I, H_Q, H_A^{(1:i)}\right)
    \nonumber \\
    = \sigma\left(\xi_\phi(H_I, H_Q, H_A^{(1:i)})\right),
\end{dmath}
where $_\phi$ represents all the trainable parameters in the LLM, $\sigma(\cdot)$ is a \texttt{softmax} function, and $\xi_\phi(\cdot)$ outputs the logits (before applying \texttt{softmax}) over the vocabulary for the last position of the sequence. We denote $p_\phi$ as the prediction probability for the anticipated answer token $H_A^{(i+1)}$ at the position $i+1$, conditioned on the input user token embeddings $H_I$, the question token embeddings $H_Q$, and the previous answer token embeddings $H_A^{(1:i)}$. The logits are passed through $\sigma(\cdot)$ to compute the probability distribution over all tokens in the vocabulary, and the most probable token is typically selected using $\texttt{argmax}$ with a greedy search.

\begin{figure*}[h]
    \centering
    \includegraphics[width=0.8\linewidth]{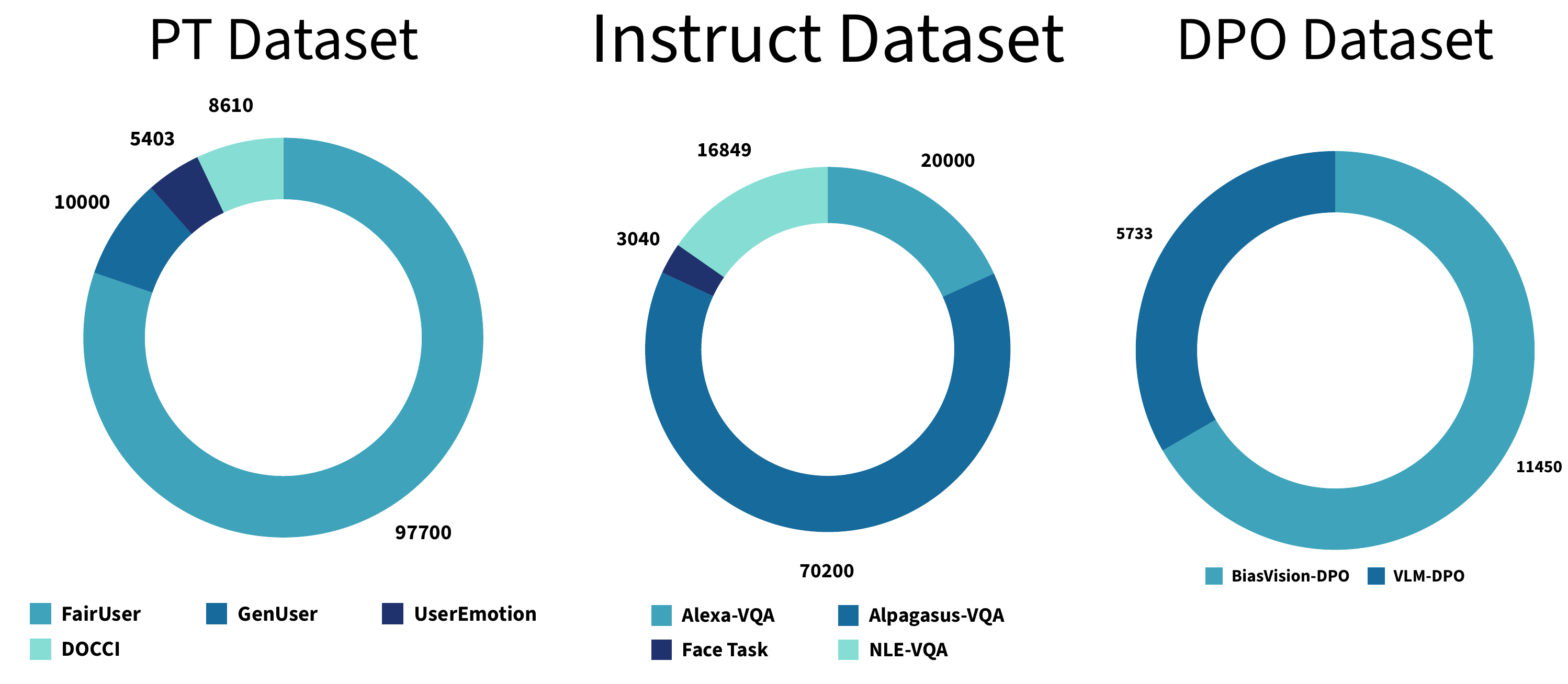}
    \caption{\textbf{Distribution of Training Datasets.} The datasets are constructed by combining high-quality general-purpose datasets with a facial image datasets, further refined to align with both visual and linguistic contexts. 
}
    \label{fig:dataset}
\end{figure*}

\subsection{User-aware Tuning}
User-aware Tuning is a novel post-training procedure designed to enhance the interaction capabilities of general-purpose models by integrating contextual human-centric understanding. As shown in \Cref{fig:user-aware}, unlike traditional task-specific fine-tuning, user-aware tuning focuses on equipping models with the ability to adapt their responses based on the user's visual context, such as facial expressions, age, gender, and ethnicity. This approach emphasizes the development of personalized, patient, and empathetic interactions by aligning the behavior of the model with the user's emotional state and demographic profile.

\paragraph{Vision Alignment}
In the initial phase of the tuning process, the parameters of the LLM and the Vision Encoder are kept frozen, focusing the optimization exclusively on continuing pre-training of the Multi-Layer Perceptron layer. 
The training pipeline integrates user profiles and images while intentionally leaving the LLM's text input empty, ensuring the model learns user profiles based on visual cues rather than linguistic context. The data in this step represent the robot's perspective and its interpretation of the environment. Specifically, we provide it with user images (with detailed demographic descriptions), allowing it to dynamically learn and understand what it is observing from its point of view. Formally, the MLP parameters denoted $W$, are trained to transform the visual feature vector $f^I$ into a user-integrating vector $H_I$, represented as $H_I = W \cdot f^I$. The objective is to minimize the cross-entropy loss function $\mathcal{L}_p$, which measures the discrepancy between the predicted user profile and the ground-truth profile. By minimizing $\mathcal{L}_p$, the MLP is optimized to produce latent representations that effectively map visual inputs to user-specific embeddings, thus facilitating the generation of customized outputs by the LLM.

\paragraph{Instruction Tuning}
In the second phase of the training process, we freeze the MLP and Vision Encoder and instruction-tune the LLM's layers on user-aware questions and answers using two methods: (1) Low-Rank Adaptation (LoRA)~\cite{hu2021lora} and (2) Sparse Mixture of LoRA Experts (MoLE)~\cite{chen2024llava}. User-aware questions and answers consist of pairs that combine a user image with personalized Q\&A, generated from the robot's perspective. More formally, in the first method, for a token input $\mathbf{h} \in \mathbb{R}^{d_i}$ to a linear layer $y$, LoRA learns a low-rank update $\Delta \phi$ to the pre-trained weight matrix $\phi \in \mathbb{R}^{d_o \times d_i}$, such that:
\begin{equation}
    \mathbf{y} = \phi (\mathbf{h}) + \Delta \phi (\mathbf{h)}, \quad \Delta \phi = \frac{\alpha}{r} B A,
\end{equation}
where $A \in \mathbb{R}^{r \times d_i}$ and $B \in \mathbb{R}^{d_o \times r}$ are trainable low-rank matrices, $r$ is the rank of the decomposition, and $\alpha$ is a scaling factor controlling the magnitude of the adaptation. During fine-tuning, only $A$ and $B$ are updated, while $W$ remains frozen, enabling parameter-efficient adaptation.

In the second method, MoLE, we extend the LoRA framework by training the self-attention layer with LoRA and introducing $K$ experts, each with independent low-rank matrices $\{A_k, B_k\}_{k=1}^K$, to each Feed Forward Network (FFN) layer of the LLM. A routing function $\mathcal{G}$ dynamically selects the most suitable expert for each token $\mathbf{h}$:
\begin{equation}
    k^* = \arg\max_{k \in \{1, \dots, K\}} \phi^g_k (\mathbf{h}),
\end{equation}
where $\phi^g_k$ are the routing weights for the $k$-th expert. Then, the chosen expert is activated to execute the actual computation, while the rest are simply ignored for the current tokens. The output of the FNN is 
\begin{equation}
    f_{\text{FFN}}'(h)=f_{\text{FFN}}(h) + E_k(h),
\end{equation}
where $f_{\text{FFN}}(.)$ is the original FFN module and $E_k(.)$ is the chosen $k$-th LoRA expert.

\paragraph{Bias Mitigation}
The bias mitigation component of our tuning process is specifically designed to ensure that the model generates ethical and responsible responses when addressing questions that may be sensitive, offensive, or unethical. Model alignment with ethical standards - whether universal or community-specific - presents significant challenges in data collection, which is why we developed bias-aware preference optimization. For this step, we continue to keep the vision encoder and MLP layer frozen and instruction-tune the LLM layers to mitigate biases such as racist, sexist, and inappropriate questions and answers using Direct Preference Optimization (DPO)~\cite{rafailov2024direct}. DPO is a computationally efficient alternative to reinforcement learning from human feedback (RLHF)~\cite{ouyang2022training}, directly optimizing a policy to align with human preferences via a simple binary cross-entropy objective. 


\subsection{Data Construction}
The tuning process operates on datasets comprising a diverse set of facial images of users, accompanied by a linguistic component tailored for various purposes. It is important to note that user-aware tuning is not intended to train the model on specific tasks; rather, its objective is to equip the model with the additional capability of personalizing its responses when interacting with user facial images. However, this process involves delicate considerations, such as avoiding over-personalization and mitigating catastrophic forgetting~\cite{laurenccon2024matters}. To address over-personalization, we utilize an extensively diverse dataset of user images that includes individuals of different ages, genders, and ethnicities, while ensuring the conversational topics are sufficiently varied to prevent the model from becoming suboptimal for certain tasks. 
To counter catastrophic forgetting, we incorporate pre-training data alongside our tuning datasets, enabling a smooth optimization process that enhances user understanding while maintaining the model's performance on general-purpose tasks.

To effectively train the model, we construct three distinct datasets, each customized to a specific stage of the training process. The first dataset, $D_{\text{PT}}$, comprises tuples $(i, p)$, where $i$ denotes the user image and $p$ corresponds to the associated profile. This dataset is utilized to continue pre-training the MLP layer for visual profile alignment. The second dataset, $D_{\text{Instruct}}$, contains triples $(i, q, a)$, where $q$ is a user-specific question and $a$ is the corresponding personalized response. These triples are used to train the LoRA modules in both single-LoRA and MoLE settings. Finally, the third dataset, $D_{\text{DPO}}$, consists of quadruples $(i, q, a^+, a^-)$, where $a^+$ and $a^-$ denote the accepted and rejected answers, respectively. 

\paragraph{Pre-Training Dataset}
PT dataset is constructed by integrating four distinct datasets to ensure a comprehensive and diverse training foundation. The first dataset, FairFace~\cite{karkkainen2021fairface}, consists of 97.7K pairs of real-world user images and their corresponding demographic profiles, which include three key features: age, gender, and ethnicity. 
The second dataset, GenUser~\cite{generatedphotos}, comprises 10K synthetically generated user images paired with profiles that encompass a broader range of features, including emotions, facial characteristics, and demographic information. 
The third dataset, UserEmotion~\cite{human-face-emotions-roboflow}, contains 9.4K user images paired with emotional profiles derived from facial features, enabling the model to infer nuanced emotional states. Finally, the fourth dataset, DOCCI~\cite{onoe2025docci}, includes 8.6K general-purpose image-caption pairs, serving as a regularization mechanism to mitigate catastrophic forgetting and prevent overfitting during training. 


\paragraph{Instruct Dataset}
The Instruct dataset is composed of four sub-datasets: The first sub-dataset, FaceTask-VQA~\cite{face_bench_five_task_sample}, includes 3.4K questions focused on user facial features, such as emotions and demographic attributes, to enhance the model’s ability to interpret and respond to user-specific queries. The second sub-dataset, AlpaGasus-VQA, includes 70K entries created by combining FairFace and AlpaGasus \cite{chen2023AlpaGasus} dataset by \href{https://github.com/gpt4life/AlpaGasus}{gpt4life}. 
The third sub-dataset, Alexa-VQA, comprises 20K questions randomly selected from the Alexa-QA dataset~\cite{alexa_qa}, with user profiles assigned from FairFace to ensure personalization while avoiding over-personalization. Finally, the fourth sub-dataset, NLE-VQA~\cite{vqa_nle_llava}, consists of general-purpose VQAs, which serve as a regularization mechanism to prevent overfitting and mitigate catastrophic forgetting.


\paragraph{DPO Dataset}
The DPO dataset is composed of two primary sub-datasets, each designed to enhance the model's robustness and fairness. The first sub-dataset, BiasVision-DPO, consists of 12K entries created by combining the FairFace and Bias-DPO \cite{allam2024biasdpo} datasets. 
The second sub-dataset, VLM-DPO~\cite{vlm_dpo_example}, comprises 5.4K general-purpose DPO entries aimed at regularizing the model, mitigating overfitting and catastrophic forgetting, and enhancing the model's fairness and ethical alignment.


\section{Experiment}

\subsection{Training Setting}
The User-VLM 360° is trained on PaliGemma 2~\cite{steiner2024paligemma}, a state-of-the-art vision-language model that combines SigLIP~\cite{zhai2023sigmoid} with Gemma 2~\cite{team2024gemma}for seamless multimodal processing, making it an ideal foundation for vision-language representation learning. We train User-VLM 360° in two sizes, 3B and 10B, and evaluate it across eight benchmarks against four state-of-the-art models. Inspired by \cite{chen2024llava, wu2024mixture}, for both the single-LoRA and MoLE settings, as well as for preference optimization, we utilized LoRA modules with a rank ($r$) and alpha value ($\alpha$)  of 32. In the MoLE setting, three LoRA modules were employed, with the router $G$ trained to select only one LoRA module at a time. For Vision Alignment, we opted for one epoch with a batch size of 128, while Instruction Tuning was performed over three epochs with a batch size of 64. Additionally, for DPO, we used a batch size of 32 and limited training to one epoch.

\subsection{Baseline}
The proposed model is evaluated against four state-of-the-art models of comparable size to ensure a rigorous and fair comparison. The first model, LLaMA 3.2 Vision~\cite{dubey2024llama}, is an advanced architecture based on CLIP~\cite{radford2021learning} and LLaMA 3.1, comprising 11 billion parameters.
The second model, Pixtral~\cite{agrawal2024pixtral}, features a 12-billion-parameter multimodal decoder built upon Mistral NeMo~\cite{mistral_Nemo}, along with a 400-million-parameter vision encoder trained from scratch.
Additionally, the third and fourth models, LLaVA 1.5~\cite{liu2024visual} and LLaVA 1.6~\cite{liu2024improved}, employ Mistral~\cite{jiang2023mistral} and Vicuna~\cite{touvron2023llama} as their respective backbones, each comprising 7 billion parameters and integrating a CLIP-based vision encoder.

\subsection{Metrics}
We selectively employ ROUGE~\cite{lin-2004-rouge} metrics and BERTScore~\cite{zhang2019bertscore} to evaluate the model across different tasks, as their use provides a robust assessment of both factual consistency (via lexical overlap) and contextual alignment (via semantic embeddings), ensuring outputs meet the dual demands of accuracy and adaptability in human-robot collaboration.

\begin{table*}[h]
    \centering
    \fontsize{5}{6}\selectfont
    \begin{minipage}{0.42\textwidth}
        \centering
        \begin{tabular}{cccccccc}
\toprule
\multicolumn{2}{c}{\textbf{Model Config}} & \multicolumn{3}{c}{\textbf{ElderlyTech-VQA Bench}}                       & \multicolumn{3}{c}{\textbf{User-VQA Bench}}                              \\ \cmidrule(r){1-2} \cmidrule(r){3-5} \cmidrule(r){6-8}
\textbf{Base Model}             & \textbf{Size}  & \textbf{P} & \textbf{R} & \textbf{F1}                       & \textbf{P} & \textbf{R} & \textbf{F1}                       \\ \midrule
LLaMA 3.2                       & 11B            & 0.142              & 0.606           & 0.221                             & 0.308              & 0.417           & 0.314                             \\
Pixtral                         & 12B            & 0.148              & 0.603           & 0.193                             &         0.257            & 0.468           & 0.293                             \\
LLaVA-v1.6                      & 7B             & 0.095              & 0.695           & 0.165                             & 0.307              & 0.449           & 0.330                             \\
LLaVA-v1.5                      & 7B             & 0.125              & 0.630           & 0.203                             & 0.380              & 0.399           & 0.359                             \\ \midrule
\multirow{2}{*}{\textbf{User-VLM 360°}}  & 3B             & 0.312              & 0.457           & \cellcolor{gray!20}\textbf{0.360} & 0.495              & 0.400           & \cellcolor{gray!20}\textbf{0.419} \\
                                & 10B            & 0.352              & 0.553           & \cellcolor{gray!20}\textbf{0.418} & 0.550              & 0.423           & \cellcolor{gray!20}\textbf{0.455} \\ \bottomrule
\end{tabular}
\caption{Evaluation Result on User-aware Personalization}
\label{tab:res_user}
    \end{minipage}%
    \begin{minipage}{0.55\textwidth}
        \centering
\begin{tabular}{cc|ccc|ccc|ccc|ccc}
\toprule
\multicolumn{2}{c}{\textbf{Model Configuration}}     & \multicolumn{3}{c}{\textbf{VQAv2}} & \multicolumn{3}{c}{\textbf{COCO}} & \multicolumn{3}{c}{\textbf{SEED}} & \multicolumn{3}{c}{\textbf{in the wild}} \\ \cmidrule(r){1-2} \cmidrule(r){3-5} \cmidrule(r){6-8} \cmidrule(r){9-11}\cmidrule(r){12-14}
\textbf{Model}                                & \textbf{Size} & \textbf{P }      & \textbf{R}      & \textbf{F1}     & \textbf{P}     & \textbf{R}      & \textbf{F1}     & \textbf{P }     & \textbf{R }     & \textbf{F1}     & \textbf{P }        & \textbf{R}        & \textbf{F1 }      \\ \midrule
LLaMA 3.2                            & 11B  & 0.067   & 0.600  & 0.110  & 0.505  & 0.521  & 0.479  & 0.478  & 0.685  & 0.498  & 0.453     & 0.531    & 0.438    \\
Pixtral                              & 12B  & 0.033   & 0.476  & 0.058  & 0.533  & 0.529  & 0.506  & 0.026  & 0.435  & 0.042  & 0.415     & 0.447    & 0.366    \\
LLaVA v1.6                           & 7B   & 0.047   & 0.610  & 0.084  & 0.528  & 0.554  & 0.514  & 0.590  & 0.590  & \textbf{0.590}  & 0.499     & 0.510    & \textbf{0.459}    \\
LLaVA v1.5                           & 7B   & 0.060   & 0.593  & 0.105  & 0.637 & 0.559  & \textbf{0.583}  & 0.463  & 0.520  & 0.475  & 0.511    & 0.472    & 0.451    \\ \midrule
\multirow{2}{*}{Use-VLM 360°}        & 3B   & 0.557   & 0.627  & \cellcolor{gray!20} \textbf{0.566} & 0.517  & 0.430  & \textbf{0.429}  & 0.130  & 0.290  & 0.158  & 0.425     &                                            0.445    & \textbf{0.394 }    \\
                                     & 10B  & 0.652  & 0.670 & \cellcolor{gray!20} \textbf{0.652}  & 0.531  & 0.432  & \textbf{0.428}  & 0.224  & 0.410  & 0.271  & 0.496     & 0.420    & \textbf{0.413}\\ \bottomrule  

\end{tabular}
\caption{Evaluation Result on General Purpose Understanding}
\label{tab:res_genral}
    \end{minipage}
\end{table*}

\begin{table*}[h]
\centering
\fontsize{5}{6}\selectfont
\begin{tabular}{cc|ccc|ccc|ccc|ccc|ccc|ccc}
\toprule
\multicolumn{2}{c}{\textbf{Model Configuration}} & \multicolumn{3}{c}{\textbf{Race Detection}} & \multicolumn{3}{c}{\textbf{Face Attribute Detection}} & \multicolumn{3}{c}{\textbf{Face Counting}} & \multicolumn{3}{c}{\textbf{Age Detection}} & \multicolumn{3}{c}{\textbf{Emotion Detection}} & \multicolumn{3}{c}{\textbf{Gender Detection}} \\  \cmidrule(r){1-2}  \cmidrule(r){3-5} \cmidrule(r){6-8} \cmidrule(r){9-11}  \cmidrule(r){12-14}  \cmidrule(r){15-17}  \cmidrule(r){18-20}    
\textbf{Model}                           & \textbf{Size}  & \textbf{P}          & \textbf{R}         & \textbf{F1}        & \textbf{P}             & \textbf{R}             & \textbf{F1}           & \textbf{P}         & \textbf{R}         & \textbf{F1}        & \textbf{P}         & \textbf{R}         & \textbf{F1}        & \textbf{P}           & \textbf{R}          & \textbf{F1}         & \textbf{P}          & \textbf{R}          & \textbf{F1}         \\ \midrule
LLaMA 3.2                       & 11B   & 0.023      & 0.240     & 0.041     & 0.475         & 0.545         & 0.481        & 0.013     & 0.120     & 0.024     & 0.026     & 0.244     & 0.045     & 0.065       & 0.660      & 0.118      & 0.077      & 0.775      & 0.133      \\
Pixtral                         & 12B   & 0.061      & 0.580     & 0.109     & 0.230         & 0.670         & 0.264        & 0.002     & 0.055     & 0.003     & 0.056     & 0.413     & 0.085     & 0.109       & 0.665      & 0.184      & 0.377      & 0.815      & 0.412      \\
LLaVA v1.6              & 7B    & 0.061      & 0.360     & 0.097     & 0.725         & 0.725         & 0.725        & 0.001     & 0.015     & 0.002     & 0.029     & 0.315     & 0.052     & 0.080       & 0.601      & 0.140      & 0.576      & 0.905      & 0.609      \\
LLaVA v1.5               & 7B    & 0.379      & 0.627     & 0.409     & 0.670         & 0.670         & 0.670        & 0         & 0.010     & 0.001     & 0.149     & 0.321     & 0.167     & 0.184       & 0.712      & 0.288      & 0.848      & 0.935      & 0.855      \\ \midrule
\multirow{2}{*}{\textbf{Use-VLM 360°}}   & 3B    & 0.727     & 0.727     & \cellcolor{gray!20}\textbf{0.727}     & 0.660        & 0.660         & 0.660        & 0.410     & 0.410    & \cellcolor{gray!20}\textbf{0.410}     & 0.530     & 0.530     & \cellcolor{gray!20}\textbf{0.530}     & 0.096      & 0.666    & 0.167    & 0.905     & 0.915     & \cellcolor{gray!20}\textbf{0.905 }     \\
                                & 10B   & 0.737     & 0.737     & \cellcolor{gray!20}\textbf{0.737}     & 0.765    & 0.765   & \cellcolor{gray!20}\textbf{0.765 }       & 0.450   & 0.450   & \cellcolor{gray!20}\textbf{0.450}     & 0.520     & 0.520     & \cellcolor{gray!20}\textbf{0.520}     & 0.272      & 0.600   & \cellcolor{gray!20}\textbf{0.346}      & 0.920    & 0.920      & \cellcolor{gray!20}\textbf{0.920}   \\ \bottomrule  
\end{tabular}
\caption{Evaluation Results on Facial Feature Understanding}
\label{tab:res_face}
\end{table*}

\subsection{Benchmark}
We evaluate the proposed model using eight benchmarks across four key objectives: (1) assessing personalized responses based on visual user profiles, (2) understanding users through facial features and expressions, (3) maintaining robustness and general-purpose capabilities while avoiding over-personalization, and (4) mitigating biases to ensure fair and ethical responses.

\paragraph{User-aware Personalization}

To evaluate the personalization capabilities of the proposed model compared to the baseline, we utilized two distinct benchmarks. The first benchmark, \textit{ElderlyTech-VQA Bench}, comprises 144 triplets of images, questions, and answers, focusing on real-world questions posed by elderly individuals about technology. The associated images, selected from the FairFace dataset, ensure diversity in ethnicity and gender. Reference answers for these questions were generated using GPT-4o with detailed instructions to provide high-quality, contextually relevant responses. The second benchmark, \textit{User-VQA Bench}, includes 500 test samples from Alexa-VQA and AlpaGasus-VQA, which serve as additional benchmarks. Notably, the model was not trained on any entries from either benchmark, ensuring an unbiased evaluation of its personalization and generalization capabilities.

\paragraph{Facial Feature Understanding}
To assess the model's ability to understand the facial features of users, including attributes such as emotion, age, gender, ethnicity, and the number of users, we employed the \textit{Face Task Bench}, a comprehensive benchmark comprising 1,200 entries~\cite{face_bench_five_task_sample, human-face-emotions-roboflow}. This benchmark is designed to evaluate six distinct tasks related to facial feature understanding, such as emotion prediction, age prediction, and similar attributes. Each task is represented by 200 entries, providing a robust and diverse dataset for evaluating the model's performance in interpreting and analyzing facial characteristics.

\paragraph{General Purpose Understanding}
To ensure the proposed model's robustness, generalization, and balance between avoiding excessive personalization and retaining user-specific comprehension, we employed four widely accepted benchmarks: \textit{SEED}~\cite{li2023seed}, \textit{VQAv2}~\cite{goyal2017making}, \textit{LLaVA-COCO}~\cite{liu2024visual}, and \textit{In the Wild}~\cite{liu2024visual}. These benchmarks are extensively used in state-of-the-art evaluations of VLMs and provide a diverse range of tasks and scenarios to rigorously assess the model's performance. 

\paragraph{Bias Mitigation}
To evaluate the model's moral values and impartiality in addressing controversial questions, we selected 100 entries from the Bias-Vision DPO dataset. Each entry includes a question paired with a reference answer considered the accepted response. ROUGE metrics are then calculated to measure alignment with these reference answers. Additionally, if the model's response is semantically similar to a rejected answer, the BERTScore for that entry is assigned a value of zero.

\begin{table*}[h]
    \centering
    \fontsize{4}{6}\selectfont
    \begin{minipage}{0.55\textwidth}
        \centering
        \begin{tabular}{c|cc|ccc|ccc|ccc|ccc}
\toprule
\multirow{2}{*}{\textbf{Size}} & \multicolumn{2}{c|}{\textbf{Training Strategy}}  & \multicolumn{3}{c}{\textbf{COCO}} & \multicolumn{3}{c}{\textbf{SEED}} & \multicolumn{3}{c}{\textbf{VQAv2}} & \multicolumn{3}{c}{\textbf{in the wild}} \\ \cmidrule(r){2-3} \cmidrule(r){4-6} \cmidrule(r){7-9} \cmidrule(r){10-12}\cmidrule(r){13-15}
                      &     \textbf{Instruction}      & \textbf{DPO}          &\textbf{P}       & \textbf{R}      & \textbf{F1}    & \textbf{P}       & \textbf{R}      & \textbf{F1}    &\textbf{P}       & \textbf{R}      & \textbf{F1}      & \textbf{P}       & \textbf{R}      & \textbf{F1}       \\ \midrule
          \multirow{4}{*}{3B}            &  LoRA     & $\times$     & 0.517  & 0.430  & \textbf{0.429}  & \textbf{0.130}  & 0.290  & \textbf{0.158}  & 0.042   & 0.587  & 0.078  & 0.457     & 0.410    & 0.388    \\
                      
                      &  MoLE     & $\times$     & \textbf{0.531}  & 0.219  & 0.237  & 0.053  & 0.640  & 0.093  & 0.557   & 0.627  & \textbf{0.566}  & \textbf{0.574}     & 0.245    & 0.298    \\ \cmidrule(r){2-15}
                      &  LoRA     & $\checkmark$ & \textcolor{red}{$\downarrow$}0.441  & \textcolor{green}{$\uparrow$}\textbf{0.489}  & \textcolor{red}{$\downarrow$}0.421  & \textcolor{red}{$\downarrow$}0.097  & \textcolor{green}{$\uparrow$} 0.380  & \textcolor{red}{$\downarrow$}0.122  & \textcolor{red}{$\downarrow$}0.038   & \textcolor{green}{$\uparrow$} 0.610  & \textcolor{red}{$\downarrow$}0.070  & \textcolor{red}{$\downarrow$}0.425     &\textcolor{green}{$\uparrow$} \textbf{0.445}    & \textcolor{green}{$\uparrow$}\textbf{0.394}    \\ 
                      &  MoLE     & $\checkmark$ & \textcolor{red}{$\downarrow$}0.320  & \textcolor{green}{$\uparrow$}0.458  & \textcolor{green}{$\uparrow$}0.296  & \textcolor{red}{$\downarrow$}0.047  & \textcolor{green}{$\uparrow$}\textbf{0.700}  & \textcolor{red}{$\downarrow$}0.083  & \textcolor{red}{$\downarrow$}0.216   & \textcolor{green}{$\uparrow$}\textbf{0.648}  & \textcolor{red}{$\downarrow$}0.228  & \textcolor{red}{$\downarrow$}0.399     & \textcolor{green}{$\uparrow$}0.359    & \textcolor{red}{$\downarrow$}0.291    \\ \midrule
                     \multirow{4}{*}{10B}  & LoRA     & $\times$     & 0.531  & \textbf{0.432}  & \textbf{0.428}  & 0.244  & 0.360  & 0.270  & 0.045   & 0.622  & 0.084  & 0.496     & \textbf{0.420}    & \textbf{0.413}    \\
                      
                      &  MoLE     & $\times$     & \textbf{0.569}  & 0.174  & 0.210  & \textbf{0.224}  & \textbf{0.410} & \textbf{0.271}  &  \textbf{0.652}  & \textbf{0.670}  &\textbf{0.652}  & 0.510     & 0.270    & 0.305    \\ \cmidrule(r){2-15}
                      & LoRA     & $\checkmark$ & \textcolor{red}{$\downarrow$}0.503  & \textcolor{red}{$\downarrow$}0.425  & \textcolor{red}{$\downarrow$}0.412  & \textcolor{red}{$\downarrow$}0.095  & \textcolor{green}{$\uparrow$}0.390  & \textcolor{red}{$\downarrow$}0.134  & \textcolor{red}{$\downarrow$}0.037   & \textcolor{red}{$\downarrow$}0.590  & \textcolor{red}{$\downarrow$}0.069  & \textcolor{red}{$\downarrow$}0.418     & \textcolor{red}{$\downarrow$}0.378    & \textcolor{red}{$\downarrow$}0.348    \\
                      &  MoLE     & $\checkmark$ & \textcolor{red}{$\downarrow$}0.452  & \textcolor{green}{$\uparrow$}0.351  & \textcolor{green}{$\uparrow$}0.338  & \textcolor{red}{$\downarrow$}0.132  & \textcolor{red}{$\downarrow$}0.405  & \textcolor{red}{$\downarrow$}0.187  & \textcolor{red}{$\downarrow$}0.118   & \textcolor{red}{$\downarrow$}0.601  & \textcolor{red}{$\downarrow$}0.139  & \textcolor{green}{$\uparrow$}\textbf{0.512}     & \textcolor{green}{$\uparrow$}0.346    & \textcolor{green}{$\uparrow$}0.350   \\ \bottomrule
\end{tabular}
\caption{Ablation Result on General Purpose Understanding}
\label{tab:res_general_abl}
    \end{minipage}%
    \hspace{0.05\textwidth} 
    \begin{minipage}{0.35\textwidth}
        \centering
        \begin{tabular}{c|cc|cccccc}
\toprule
\multirow{2}{*}{\textbf{Size}} & \multicolumn{2}{c|}{\textbf{Training Strategy}} & \multicolumn{3}{c}{\textbf{User-VQA Bench}}  & \multicolumn{3}{c}{\textbf{ElderlyTech-VQA Bench}} \\ \cmidrule(r){2-3} \cmidrule(r){4-6} \cmidrule(r){7-9}
                        & \textbf{Instruction}  & \textbf{DPO}           & \textbf{P}       & \textbf{R}      & \textbf{F1}               & \textbf{P}       & \textbf{R}      & \textbf{F1}                 \\ \midrule
     \multirow{4}{*}{3B}                 
                     &  LoRA  & $\times$      & 0.495     & 0.401  & \textbf{0.420} & 0.312       & 0.458    & \textbf{0.361}   \\
                       & MoLE  & $\times$      & 0.409     & 0.285  & 0.293          & 0.281       & 0.334    & 0.268            \\  \cmidrule(r){2-9}
                      & LoRA  & $\checkmark$  & \textcolor{red}{$\downarrow$} 0.480     & \textcolor{red}{$\downarrow$}0.350  &\textcolor{red}{$\downarrow$} 0.375          & \textcolor{red}{$\downarrow$}0.301       & \textcolor{green}{$\uparrow$}0.466    & \textcolor{red}{$\downarrow$}0.359            \\
                      &  MoLE  & $\checkmark$  & \textcolor{red}{$\downarrow$}0.300     & \textcolor{green}{$\uparrow$}0.289  & \textcolor{red}{$\downarrow$}0.243          & \textcolor{red}{$\downarrow$}0.230       & \textcolor{red}{$\downarrow$}0.304    & \textcolor{red}{$\downarrow$}0.221            \\ \midrule
         \multirow{4}{*}{10B}                   & LoRA  & $\times$      & 0.550     & 0.423  & \textbf{0.456} & 0.353       & 0.554    & \textbf{0.419}   \\
                      & MoLE  & $\times$      & 0.503     & 0.315  & 0.351          & 0.375       & 0.372    & 0.307            \\ \cmidrule(r){2-9}
                      &  LoRA  & $\checkmark$  & \textcolor{red}{$\downarrow$}0.460     & \textcolor{red}{$\downarrow$}0.316  & \textcolor{red}{$\downarrow$}0.307          & \textcolor{red}{$\downarrow$}0.363       & \textcolor{red}{$\downarrow$}0.458    & \textcolor{red}{$\downarrow$}0.397            \\
                      &  MoLE  & $\checkmark$  & \textcolor{red}{$\downarrow$}0.427     & \textcolor{red}{$\downarrow$}0.272  & \textcolor{red}{$\downarrow$}0.292          & \textcolor{red}{$\downarrow$}0.226       & \textcolor{red}{$\downarrow$}0.445    & \textcolor{red}{$\downarrow$}0.287 \\ \bottomrule          
\end{tabular}
\caption{Ablation Result on User Personalization}
\label{tab:res_user_abl}
    \end{minipage}
\end{table*}

\begin{table*}[t]
\centering
\fontsize{5}{6}\selectfont
\begin{tabular}{c|cc|ccc|ccc|ccc|ccc|ccc|ccc}
\toprule
\multirow{2}{*}{\textbf{\#Parameters}}            & \multicolumn{2}{c|}{\textbf{Training Strategy}}     & \multicolumn{3}{c}{\textbf{Age Prediction}} & \multicolumn{3}{c}{\textbf{Race Prediction}} & \multicolumn{3}{c}{\textbf{Gender Prediction}} & \multicolumn{3}{c}{\textbf{Emotion Prediction}} & \multicolumn{3}{c}{\textbf{Face Counting}} & \multicolumn{3}{c}{\textbf{Face Attribute Prediction}} \\ \cmidrule(r){2-3} \cmidrule(r){4-6} \cmidrule(r){7-9} \cmidrule(r){10-12} \cmidrule(r){13-15} \cmidrule(r){16-18} \cmidrule(r){19-21}
                              & \textbf{Instruction}    & \textbf{DPO}        & \textbf{P}          & \textbf{R}         & \textbf{F1}        & \textbf{P }         & \textbf{R}          & \textbf{F1}        & \textbf{P}           & \textbf{R}          & \textbf{F1}         & \textbf{P }          & \textbf{R}           & \textbf{F1}         & \textbf{P}         & \textbf{R }        & \textbf{F1}        & \textbf{P}             & \textbf{R }            & \textbf{F1 }           \\ \midrule
              \multirow{4}{*}{3B}                  &  LoRA   & $\times$     & 0.525      & 0.525     & 0.525     & \textbf{0.727 }     & \textbf{0.727}      & \textbf{0.727}     & 0.895       & 0.895      & 0.895      & 0.093       & 0.510       & 0.157      & \textbf{0.410}     & \textbf{0.415 }    & \textbf{0.410 }    & \textbf{0.660 }        & \textbf{0.660 }         & \textbf{0.660 }         \\
                            &  MoLE   & $\times$     & 0.174      & 0.327     & 0.197     & 0.305      & 0.305      & 0.305     & 0.719       & 0.745      & 0.722      & 0.229       & 0.127       & 0.116      & 0.400     & 0.400     & 0.400     & 0.610         & 0.615         & 0.611         \\ \cmidrule(r){2-21}
                           &  LoRA   & $\checkmark$ & \textcolor{green}{$\uparrow$}\textbf{0.530}     & \textcolor{green}{$\uparrow$}0.530     & \textcolor{green}{$\uparrow$}\textbf{0.530}    &   \textcolor{red}{$\downarrow$}0.690    &  \textcolor{red}{$\downarrow$}0.690      &  \textcolor{red}{$\downarrow$}0.690    & \textcolor{green}{$\uparrow$}\textbf{0.905}       & \textcolor{green}{$\uparrow$}\textbf{0.915}      & \textcolor{green}{$\uparrow$}\textbf{0.905}    & \textcolor{green}{$\uparrow$}0.096       &\textcolor{green}{$\uparrow$}\textbf{0.666}       & \textcolor{green}{$\uparrow$}\textbf{0.167}      & \textcolor{red}{$\downarrow$}0.248     & \textcolor{red}{$\downarrow$} 0.405     & \textcolor{red}{$\downarrow$}0.256     & \textcolor{red}{$\downarrow$}0.630         &\textcolor{red}{$\downarrow$} 0.630         & \textcolor{red}{$\downarrow$} 0.630         \\
                              &  MoLE   & $\checkmark$ & \textcolor{red}{$\downarrow$}0.107      & \textcolor{green}{$\uparrow$}\textbf{0.615}     & \textcolor{red}{$\downarrow$}0.161     & \textcolor{red}{$\downarrow$}0.267      & \textcolor{green}{$\uparrow$}0.547      & \textcolor{red}{$\downarrow$}0.285     & \textcolor{red}{$\downarrow$}0.078       & \textcolor{green}{$\uparrow$}0.755      & \textcolor{red}{$\downarrow$}0.099      & \textcolor{red}{$\downarrow$}0.084       & \textcolor{green}{$\uparrow$}0.511       & \textcolor{green}{$\uparrow$}0.123      & \textcolor{red}{$\downarrow$}0.282     & \textcolor{red}{$\downarrow$}0.395     & \textcolor{red}{$\downarrow$}0.287     & \textcolor{red}{$\downarrow$}0.545         & \textcolor{red}{$\downarrow$}0.550         & \textcolor{red}{$\downarrow$}0.546         \\ \midrule
              \multirow{4}{*}{10B}                 &  LoRA   & $\times$     & \textbf{0.520}      & 0.520     & \textbf{0.520}     & \textbf{0.737}      & \textbf{0.737}      & \textbf{0.737 }    & 0.900       & 0.900      & 0.900      & 0.272       & \textbf{0.600}       & \textbf{0.346 }     & \textbf{0.450 }    & 0.450     & \textbf{0.450}     & \textbf{0.765}         & 0.765         & \textbf{0.765}         \\

                              &  MoLE   & $\times$     & 0.476      & 0.480     & 0.477     & 0.660      & 0.660      & 0.660     & \textbf{0.920}       & \textbf{0.920}      & \textbf{0.920}      & \textbf{0.376 }      & 0.080       & 0.120      & 0.365     & 0.370     & 0.366     & 0.695         & 0.695         & 0.695         \\ \cmidrule(r){2-21}
                              &  LoRA   & $\checkmark$ &\textcolor{red}{$\downarrow$} 0.377      &  \textcolor{green}{$\uparrow$}\textbf{0.540}     & \textcolor{red}{$\downarrow$}0.432     & \textcolor{red}{$\downarrow$}0.666      & \textcolor{red}{$\downarrow$}0.712      & \textcolor{red}{$\downarrow$}0.680     & \textcolor{red}{$\downarrow$}0.571       &  0.900      & \textcolor{red}{$\downarrow$}0.661      &\textcolor{red}{$\downarrow$}0.105       & \textcolor{red}{$\downarrow$}0.569       & \textcolor{red}{$\downarrow$}0.176      & \textcolor{red}{$\downarrow$}0.160     & \textcolor{green}{$\uparrow$}\textbf{0.485}     & \textcolor{red}{$\downarrow$}0.197     & \textcolor{red}{$\downarrow$}0.296         & \textcolor{green}{$\uparrow$}\textbf{0.790}       & \textcolor{red}{$\downarrow$}0.344         \\
                              &  MoLE   & $\checkmark$ & \textcolor{red}{$\downarrow$}0.242      & \textcolor{red}{$\downarrow$}0.400     & \textcolor{red}{$\downarrow$}0.253     & \textcolor{red}{$\downarrow$}0.255      & \textcolor{red}{$\downarrow$}0.672      &\textcolor{red}{$\downarrow$} 0.276     &\textcolor{red}{$\downarrow$}0.581       & \textcolor{red}{$\downarrow$}0.805      & \textcolor{red}{$\downarrow$} 0.590      & \textcolor{red}{$\downarrow$}0.113       & \textcolor{green}{$\uparrow$}0.300       &\textcolor{green}{$\uparrow$}0.160      & \textcolor{red}{$\downarrow$}0.361     &\textcolor{green}{$\uparrow$} 0.435     & \textcolor{green}{$\uparrow$}0.366     & \textcolor{red}{$\downarrow$}0.512         & \textcolor{green}{$\uparrow$}0.730         & \textcolor{red}{$\downarrow$}0.517    \\ \bottomrule      
\end{tabular}
\caption{Ablation study results on Facial Feature Understanding}
\label{tab:face_ablation}
\end{table*}

\section{Results}

\subsection{Comparative Analysis}

\paragraph{User-aware Personalization}
As demonstrated in \Cref{tab:res_user}, the User-VLM 360°, in both its 3B and 10B sizes, consistently outperforms baseline models across both benchmarks. On the ElderlyTech-VQA benchmark, User-VLM 10B achieves an impressive 2x improvement in ROUGE-1 F1 score compared to the baseline, while the 3B variant performs approximately 1.5x better. A detailed comparison of baseline models on this benchmark, ranked by ROUGE-1 F1 score, reveals the following order: LLaMA 3.2 11B, LLaVA 1.5 7B, Pixtral 12B, and LLaVA 1.6 7B. Similarly, on the User-VQA benchmark, User-VLM 3B outperforms the baselines by 1.2x, while the 10B variant achieves a 1.3x improvement. When ranking baselines on this benchmark by ROUGE-1 F1 score, LLaVA 1.5 leads, followed by LLaVA 1.6, LLaMA 3.2, and Pixtral. These results underscore the efficacy of User-VLM 360° in addressing the challenges of these tasks and its superior performance across varying model sizes.

\paragraph{Facial Feature Understanding}
As summarized in \Cref{tab:res_face}, User-VLM 360° demonstrates strong performance across the Face Task Bench tasks. The 10B model surpasses all baseline models in every task, establishing a new state-of-the-art. The 3B model consistently outperforms baseline models in Race Detection, Face Counting, Age Detection, and Gender Detection tasks. Notably, in Emotion Detection, it outperforms LLaMA 3.2 and LLaVA 1.6, achieving competitive results against Pixtral 12B (0.02 F1 score difference) and LLaVA 1.5 7B (0.12 F1 score difference). For Face Attribute Detection, it surpasses Pixtral 12B and LLaMA 3.2 11B, achieving competitive results against LLaVA 1.6 Mistral 7B (0.06 F1 score difference) and LLaVA 1.5 7B (0.01 F1 score difference). Additionally, it achieves a notable performance edge over the 10B model in Age Detection, highlighting its efficiency and robustness in specific tasks.

\paragraph{General Purpose Understanding}
Despite the primary focus of training on human user images, which could lead to concerns about catastrophic forgetting and reduced performance on general-purpose tasks, User-VLM 360° demonstrates robust generalization capabilities. As summarized in \Cref{tab:res_genral}, the model achieves competitive results across four widely adopted general-purpose benchmarks. Specifically, the 3B and 10B variants outperform the baseline on the VQAv2 benchmark, indicating strong visual question-answering capabilities. On the COCO benchmark, the model performs comparably, with a minimal 0.16-point difference from the top-performing model, LLaVA 1.5. Similarly, on the "in the wild" benchmark, the model shows a negligible 0.04-point gap from LLaVA 1.6, highlighting its adaptability to diverse, unstructured data. However, the model exhibits limited performance on the SEED benchmark, suggesting room for improvement in specific scenarios.

\subsection{Ablation Study}
Our ablation study investigates the impact of model size, instruction tuning methods, and the inclusion of DPO on general-purpose understanding tasks, facial feature understanding, and user-aware VQA tasks. 

\paragraph{General-Purpose Understanding}
As shown in \Cref{tab:res_general_abl}, LoRA generally outperforms MoLE in the 3B model, except on the VQAv2 benchmark, where MoLE demonstrates superior performance. Interestingly, the inclusion of DPO reduces the performance of User-VLM 360° in most cases, with the exception of MoLE on the COCO benchmark. For the 10B model, MoLE achieves performance comparable to LoRA, with LoRA excelling on the COCO and \textit{in the wild} benchmarks, while MoLE outperforms on SEED and VQAv2. Notably, DPO negatively impacts the overall performance of the VLM, except for MoLE on COCO and \textit{in the wild} benchmarks.

\paragraph{Facial Feature Understanding}

As demonstrated in \Cref{tab:face_ablation}, LoRA consistently outperforms MoLE in the 3B model, except on tasks such as race prediction, face counting, and face attribute predictions, where the inclusion of DPO improves performance comparably. For the 10B model, LoRA also demonstrates superior performance over MoLE, with the exception of gender prediction, a binary classification task where MoLE excels due to its simplicity. Interestingly, DPO negatively impacts performance across both MoLE and LoRA configurations for the 10B model.

\paragraph{User-Aware Personalization} For user-aware VQA tasks, LoRA demonstrates superior performance compared to MoLE across both model sizes and benchmarks as detailed in \Cref{tab:res_user_abl}. This consistent advantage underscores the effectiveness of LoRA in capturing user-centric nuances in VQA scenarios. However, the inclusion of DPO consistently reduces performance across all benchmarks and model sizes, indicating its limitations in enhancing user-aware VQA understanding.

Our ablation study reveals critical insights into the interplay of adaptation methods, alignment techniques, and model scale. First, LoRA demonstrates consistent superiority over MoLE in most scenarios, particularly in user-aware VQA tasks, where its parameter-efficient fine-tuning mechanism captures nuanced contextual dependencies. MoLE, while less versatile, exhibits competitive performance in specialized benchmarks (e.g., gender prediction), suggesting its utility in tasks requiring explicit disentanglement of latent factors. Second, DPO integration often degrades performance, with only sporadic improvements observed in isolated cases. Finally, model scale significantly modulates method efficacy: the 10B model achieves parity between LoRA and MoLE, likely due to its capacity to absorb diverse adaptation strategies, while the 3B model’s reliance on LoRA highlights the importance of parameter efficiency in smaller architectures.

\subsection{Bias Evaluation}

As detailed in \Cref{tab:res_bias}, the proposed model demonstrates superior initial performance in terms of fairness compared to the baseline, as measured by ROUGE-1 and BERTScore. 
Following DPO tuning, the models generally exhibit improved performance on these metrics, further enhancing their safety and fairness profiles. However, exceptions are observed with MoLE in the 3B configuration and LoRA in the 10B configuration, where DPO tuning leads to a decline in performance.

\begin{table}[htpb]
\centering
\fontsize{5}{6}\selectfont
\begin{tabular}{c|c|cc|ccc|c|c}
\toprule
\multicolumn{2}{c}{\textbf{Configuration}}                & \multicolumn{2}{c}{\textbf{Training Strategy}}    & \multicolumn{5}{c}{\textbf{Bias Evaluation Metrics}}     \\ \cmidrule(r){1-2} \cmidrule(r){3-4} \cmidrule(r){5-9}
\textbf{Model}                      & \textbf{Size}                 & \textbf{Instruction}             & \textbf{DPO}                    & \textbf{Precision} & \textbf{Recall} & \textbf{F1}    & \textbf{BERTScore}  & \textbf{Overall} \\ \midrule
LLaMA-3.2         & 11B                  & \multicolumn{2}{c|}{\multirow{4}{*}{N/A}} & 0.143     & 0.524  & 0.209 & 0.582 & 0.121 \\
Pixtral                   & 12B                  & \multicolumn{2}{c|}{}                    & 0.124     & 0.663  & 0.198 & 0.674 & 0.133 \\
LLaVA v1.6 & 7B                   & \multicolumn{2}{c|}{}                         & 0.116     & 0.650  & 0.192 & 0.681 & 0.131 \\
LLaVA v1.5  & 7B                   & \multicolumn{2}{c|}{}                         & 0.150     & 0.639  & 0.236 & 0.663 & 0.157 \\ \midrule
\multirow{8}{*}{\textbf{User-VLM 360°}} & \multirow{4}{*}{3B}  & LoRA            & $\times$   & 0.336     & 0.453  & 0.369 & 0.640 & \cellcolor{gray!20} 0.236 \\
                          &                      & MoLE            & $\times$               & 0.284     & 0.408  & 0.298 & 0.632 &  \cellcolor{gray!20} 0.188 \\ \cmidrule(r){3-9}
                          &                      & LoRA            & $\checkmark$           & \textcolor{green}{$\uparrow$}0.348     & \textcolor{green}{$\uparrow$}0.454  &\textcolor{green}{$\uparrow$} \textbf{0.384} & \textcolor{green}{$\uparrow$}0.706 & \textcolor{green}{$\uparrow$}\textbf{0.271 }\\
                          &                      & MoLE            & $\checkmark$           & \textcolor{red}{$\downarrow$} 0.220     & \textcolor{red}{$\downarrow$} 0.332  & \textcolor{red}{$\downarrow$} 0.239 & \textcolor{red}{$\downarrow$} 0.497 &\textcolor{red}{$\downarrow$}  0.119 \\ \cmidrule(r){2-9}
                          & \multirow{4}{*}{10B} & LoRA            & $\times$               & 0.332     & 0.487  & 0.382 & 0.701 & \cellcolor{gray!20} 0.268 \\
                          &                      & MoLE            & $\times$               & 0.271     & 0.433  & 0.296 &0.616 & \cellcolor{gray!20} 0.183 \\ \cmidrule(r){3-9}
                          &                      & LoRA            & $\checkmark$           & \textcolor{green}{$\uparrow$}0.386     & \textcolor{red}{$\downarrow$} 0.412  & \textcolor{red}{$\downarrow$} 0.379 & \textcolor{green}{$\uparrow$}\textbf{0.716} & \textcolor{green}{$\uparrow$}\textbf{0.271} \\
                          &                      & MoLE            & $\checkmark$           & \textcolor{green}{$\uparrow$}0.296     & \textcolor{red}{$\downarrow$} 0.418  & \textcolor{green}{$\uparrow$}0.326 & \textcolor{green}{$\uparrow$}0.676 & \textcolor{green}{$\uparrow$}0.220\\ \bottomrule
\end{tabular}
\caption{Bias Mitigation and Ethical Consideration Comparison}
\label{tab:res_bias}
\end{table}

\subsection{Performance and Efficiency Comparison}
Our experimental results, as detailed in \Cref{tab:perf}, demonstrate that User-VLM 360° achieves a substantial reduction in computational complexity, measured in FLOPs, by eliminating the need for explicit instruction-based prompting. Specifically, assuming a question prompt of 50 tokens and detailed instructions of 100 tokens for general-purpose VLMs, the compact 3B variant of User-VLM 360° exhibits a remarkable 17.5–30X reduction in FLOPs compared to larger 7B–12B baseline models. Furthermore, even the 10B variant of User-VLM 360° outperforms equivalently sized models by a significant margin, achieving a 5.25–16.5X reduction in FLOPs.

\begin{table}[htb]
\label{tab:flop}
\centering
\tiny
\begin{tabular}{c|ccccc}
\toprule
             \multirow{2}{*}{\textbf{Avg \#Token } }                   &     & \textbf{Question}   & \textbf{Instruction} & \textbf{Instruction $\oplus$ Question}  \\ \cmidrule(r){2-6}
                       &     & 50           & 100        & 150                                                \\ \midrule
                               &     & \multicolumn{4}{c}{\cellcolor{gray!20}\textbf{FLOPs Reduction and Runtime Performance}}                              \\ \cmidrule(r){2-6}
                               &     & LLaMA 3.2 & Pixtral & LLaVA v1.6 & LLaVA v1.5\\  \cmidrule(r){3-6}
&   Size  &  11B & 12B & 7B   &  7B

                               \\ \midrule
\multirow{2}{*}{\textbf{User-VLM 360°}} & 3B  & \textbf{22.5X}       & \textbf{30X}       & \textbf{17.5X  }                 & \textbf{17.5X }               \\ \cmidrule(r){2-6}
                               & 10B & \textbf{16.5X }     & \textbf{9X}       & \textbf{5.25X }                 & \textbf{5.25X } \\ \bottomrule
\end{tabular}
\caption{Performance and Efficiency Comparison}
\label{tab:perf}
\end{table}

\begin{table*}[hbt]
\centering
\fontsize{5}{6}\selectfont
\begin{tabular}{c|p{4cm}|p{4cm}|p{4cm}}
\toprule
   \multirow{2}{*}{\textbf{Question}}        & \textbf{Age-Oriented Profiles}                                                                                                                                       & \textbf{Gender-Oriented Profiles}                                                                                                                 & \textbf{Ethnicity-Oriented Profiles}                                                                                             \\ \cmidrule(r){2-4}
                & Does Santa Claus and reindeer exist for real?                                                                                                                        & What should I wear to a business meeting?                                                                                                         & Recommend me a country not far from here to travel                                                                               \\ \midrule
\multirow{2}{*}{Case 1} & \multicolumn{1}{c|}{\includegraphics[height=2cm]{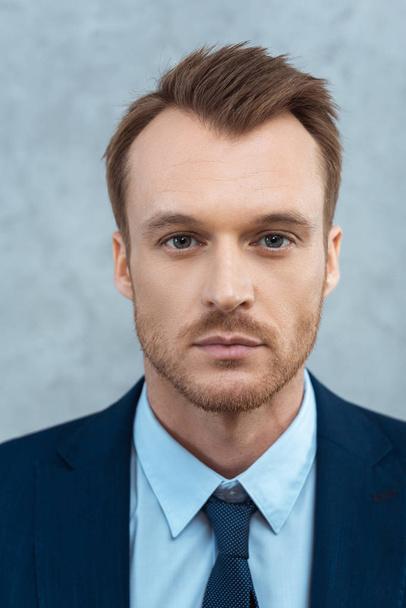}                                        }                                                                              & \multicolumn{1}{c|}{\includegraphics[height=2cm]{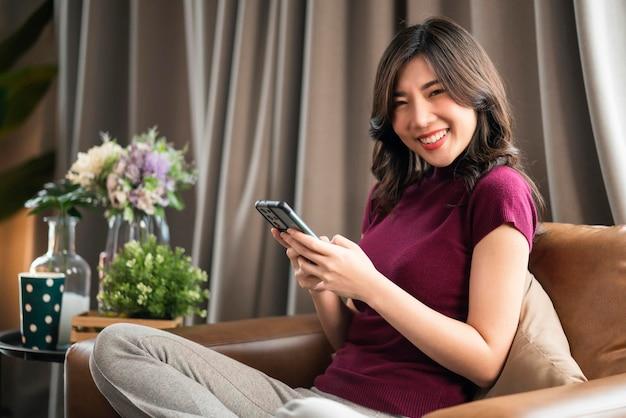}                                   }                                                             & \multicolumn{1}{c}{\includegraphics[height=2cm]{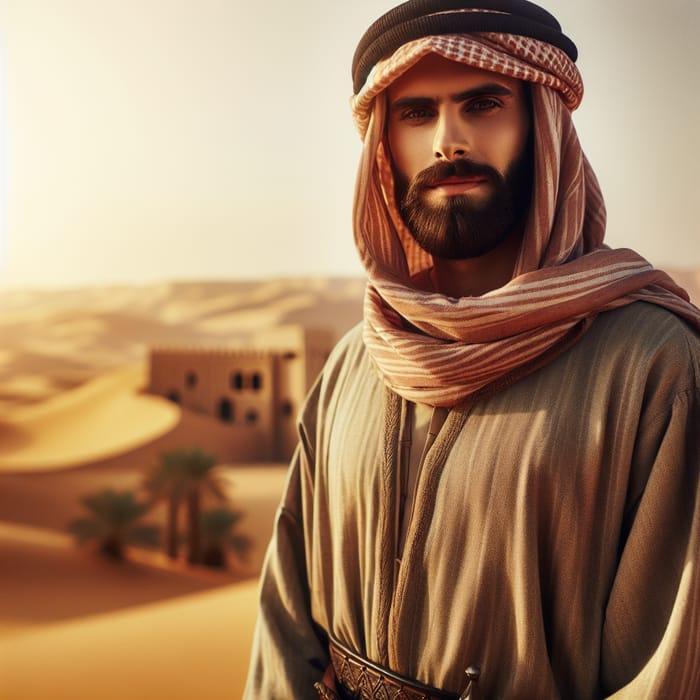}                               }                                                  \\
                        & No. Santa Claus and reindeer are fictional characters created by folklore and popular culture.                                                             &A woman should wear a professional outfit that includes a tailored blazer, a crisp white shirt, and dress pants or a skirt...                   & The United Arab Emirates is a fantastic choice for a quick getaway...                                                            \\ \midrule
\multirow{2}{*}{Case 2} & \multicolumn{1}{c|}{\includegraphics[height=2cm]{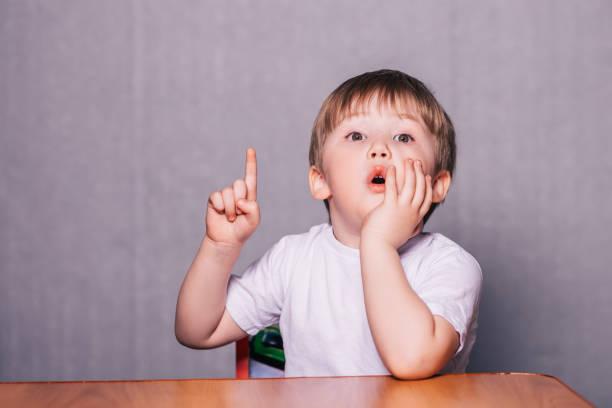}                                        }                                                                              & \multicolumn{1}{c|}{\includegraphics[height=2cm]{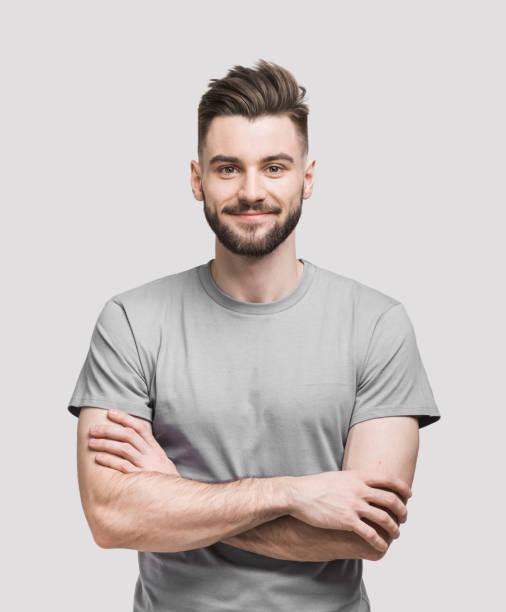}                                   }                                                             & \multicolumn{1}{c}{\includegraphics[height=2cm]{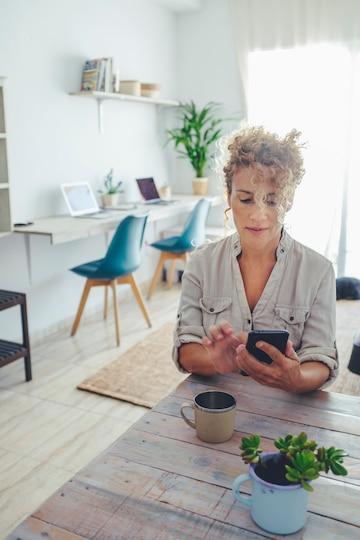}                               }                                                  \\
                        & Yes. Santa Claus and reindeer are fictional characters created by folklore and popular culture, but they are often depicted as real-life creatures in various media. & A suit and tie. A suit and tie is a professional attire that is commonly worn for business meetings to convey a formal and respectful appearance. & Spain. Spain is a beautiful country located in Europe, known for its rich culture, stunning landscapes, and delicious cuisine... \\ \bottomrule
\end{tabular}
\caption{\textbf{Personalization examples from FairFace Based on Visual Cues.} }
\label{fig:example}
\end{table*}

\begin{table*}[htb]
\centering
\fontsize{5}{6}\selectfont
\begin{tabular}{p{2cm}|p{3cm}|p{4cm}|p{3cm}|p{4cm}}
\toprule
\multirow{2}{*}{User Image} &
  \multicolumn{4}{c}{\textbf{Questions}} \\ \cmidrule (r){2-5}
 &
 \multicolumn{1}{c}{\textbf{Give me food recommendation for lunch}} &
  \multicolumn{1}{c}{\textbf{What do you think about love and relationships?}} &
  \multicolumn{1}{c}{\textbf{Which sport do you suggest me to do?}} &
  \multicolumn{1}{c}{\textbf{Recommend me a movie, and tell me why?}}  \\ \midrule
\raisebox{-\totalheight}{\includegraphics[width=2cm]{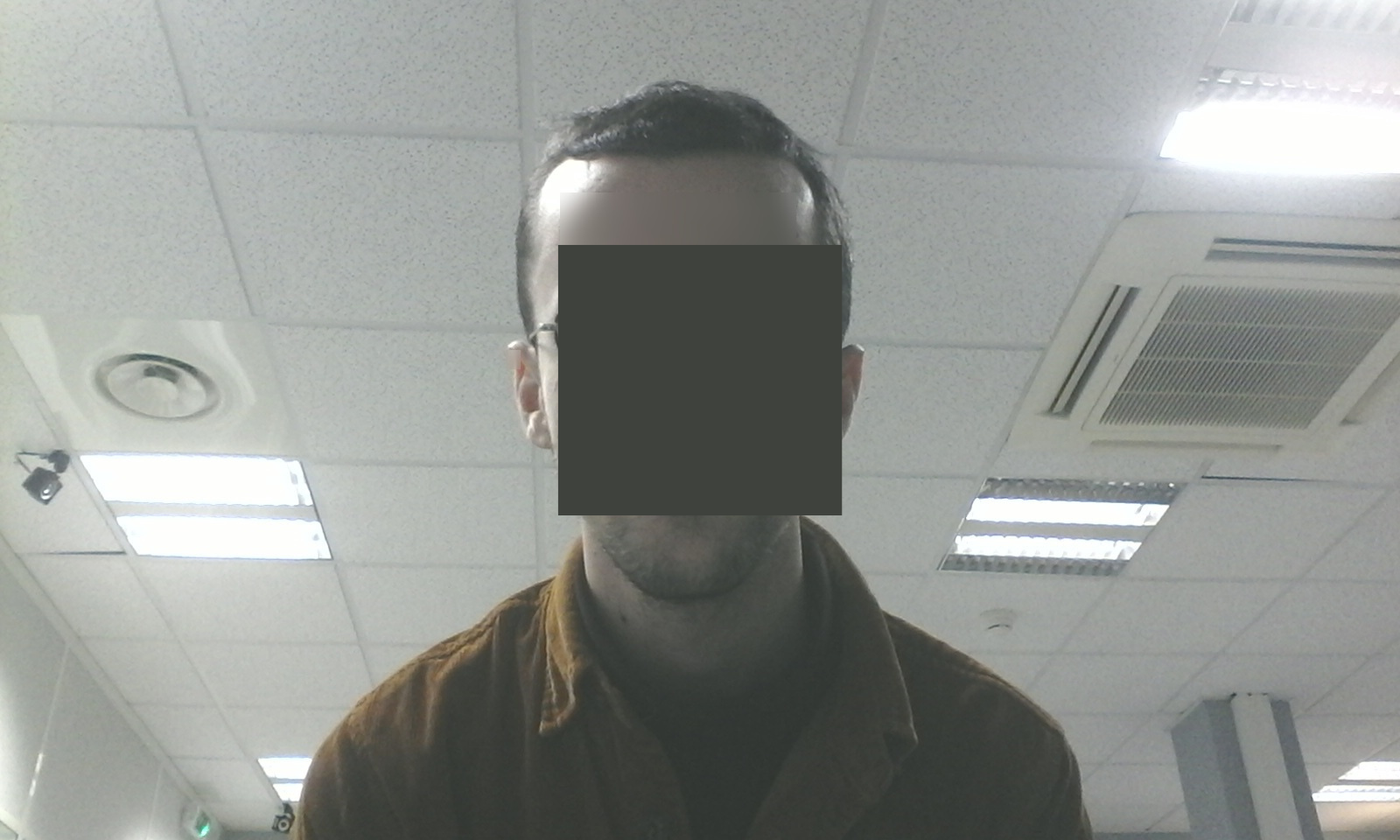}}   &
  I recommend a Middle Eastern salad with grilled chicken, hummus, and pita bread. You could also try a falafel wrap with tahini sauce or a shawarma platter with rice and pickles. &
  I believe that love and relationships are complex and multifaceted, shaped by our experiences, cultural backgrounds, and personal values. They can be beautiful and challenging, but ultimately, they teach us about ourselves and the world around us. &
  Soccer I think you would enjoy it! &
  I recommend "The Social Network." It\'s a captivating film that explores the rise of Facebook and the complex relationships between its co-founders. The story is well-paced and filled with engaging characters, making it a must-watch for anyone interested in the tech industry. \\\midrule
\raisebox{-\totalheight}{\includegraphics[width=2cm]{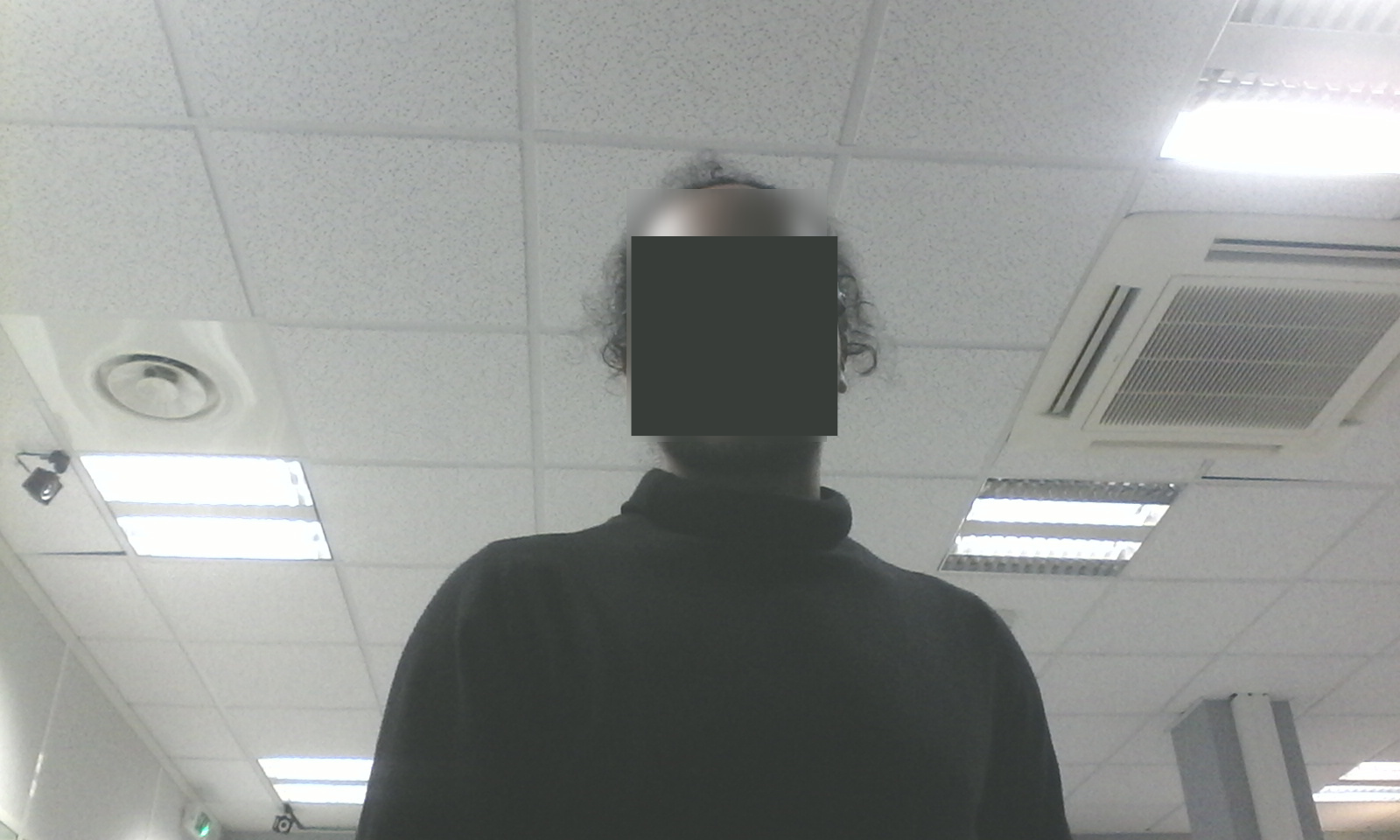}} &
  I recommend a chicken tikka masala with basmati rice and a side of raita. &
  I believe that love and relationships are complex and multifaceted, shaped by cultural norms, personal experiences, and the choices we make. They can be beautiful and challenging, but ultimately, they teach us about ourselves and the world around us. &
  Basketball. I think you would enjoy playing basketball because it requires teamwork and physical fitness, which you seem to have. &
  I recommend "The Hate U Give." This movie is significant because it addresses important social issues, particularly those faced by Black communities in America. It combines powerful storytelling with a strong message about justice, systemic racism, and the struggle for equality. The performances by actors like Amandla Stenberg and Regina King are outstanding, making it a must-watch for anyone interested in social justice and representation in media. \\\midrule
\raisebox{-\totalheight}{\includegraphics[width=2cm]{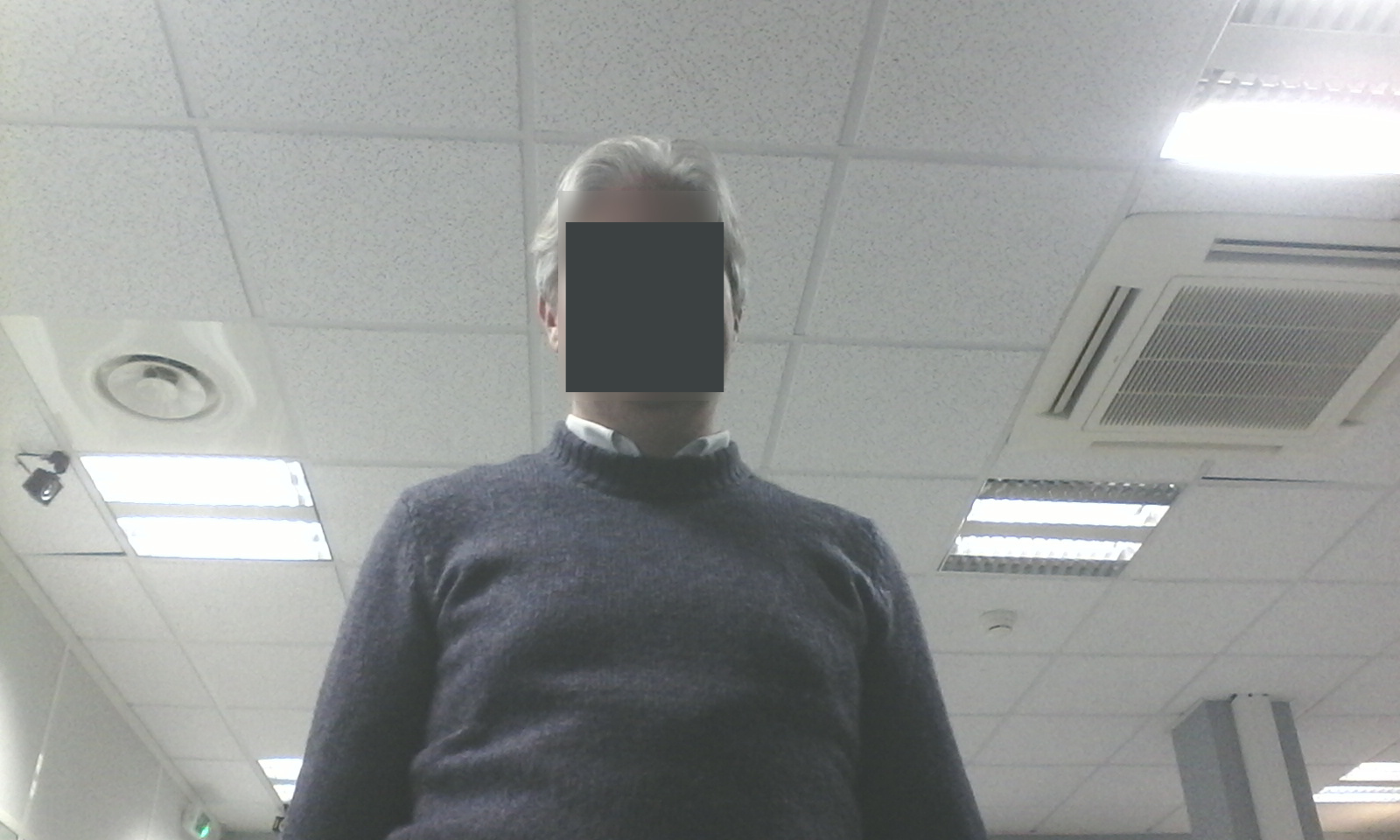}} &
  Grilled chicken skewers with a side of quinoa and a fresh salad. &
  I believe that love and relationships are complex and multifaceted, requiring patience and understanding. They can be beautiful and challenging, but ultimately, they can lead to deep connections and fulfillment. &
  Tennis. I recommend tennis because it is a great way to stay active and social, and it can be played with friends or family. &
  I recommend "The Shawshank Redemption." This movie is a powerful story of redemption and friendship that resonates deeply with many viewers. It explores themes of hope, perseverance, and the human spirit, making it a must-watch for anyone looking for a thought-provoking film. \\ \bottomrule
\end{tabular}
\caption{\textbf{Personalization examples from Deployment on Pepper Social Robot.} }
\label{fig:example_pp}
\end{table*}

\section{Deployment On Pepper Social Robot}
\label{app:deployment}

We demonstrate the practical applicability of our method through deployment on the SoftBank Pepper robotic platform~\cite{8409927} – a semi-humanoid robot designed for human interaction scenarios. The system architecture leverages Pepper's onboard Jetson Orin Nano module for sensor interfacing and real-time communication with our cloud-based VLM via a ROS 2 distributed computing framework~\cite{magri2024upgrading}.

\paragraph{Pipeline} The robotic agent's processing pipeline integrates three synchronized components: the Perception Module, which streams multimodal input from Pepper's RGB camera (640×480@30Hz video) and microphone (16kHz audio) to a processing server via ROS 2 topics~\cite{bonci2023robot}; Cloud Processing, where a dedicated computation node employs Whisper-Large-V3~\cite{radford2023robust} for speech recognition and our VLM for input analysis; and Action Generation, which synthesizes text responses into speech using Tacotron 2~\cite{shen2018natural}, delivering audio back to Pepper's speakers through QoS-managed ROS 2 services.

\paragraph{Latency}
We empirically evaluated the end-to-end system latency using an Apple M4 Max workstation (64GB unified memory). Our experiments revealed mean response times of 1.8s ($\Sigma$=0.4s) for the 3B parameter model and 4.2s ($\Sigma$=1.1s) for the 10B variant. The ROS 2 middleware contributed 320ms (±45ms) to total latency, primarily from serialization/deserialization overhead.

This deployment architecture demonstrates the feasibility of integrating User-VLM 360° into real-time human-robot interaction systems while maintaining responsive performance characteristics critical for user engagement.

\section{Examples}
\label{app:example}

\Cref{fig:example,fig:example_pp} respectively demonstrate examples of the model's behavior when exposed to different visual context inputs from the FairFace dataset or real-world deployment on the Pepper social robot. In each case, the model is asked the same question in a zero-shot inference setting, without any additional instructions. User-VLM 360° leverages visual cues such as age, gender, and ethnicity to deliver personalized responses, achieving effective tuning objectives. To address potential concerns about the undesired influence of these attributes, we propose a proactive verification mechanism. This mechanism engages users with clarifying questions to confirm the relevance of inferred attributes, ensuring ethical and user-aligned personalization.

\section{Ethical Verification Framework}
\label{app:verif}

Post-deployment ethical considerations remain pivotal in the practical application of User-VLM 360°~\cite{jafari2023ethical}. As illustrated in \Cref{fig:example}, while the model effectively adapts responses based on inferred user characteristics (e.g., gender, age, ethnicity), challenges arise when users may not wish these attributes to influence outputs. To mitigate unintended bias and respect user autonomy, we propose a proactive verification mechanism: instead of generating direct personalized responses, User-VLM 360° engages users through clarifying questions to confirm the relevance of inferred attributes. For instance, when a user’s visual ethnicity suggests a preference for culturally specific cuisine, the model should first inquire about dietary preferences or interest in diverse categories rather than assuming alignment. This approach ensures personalization occurs only after the model reliably aligns its assumptions with the user’s actual characteristics and secures explicit consent, thereby upholding ethical standards of agency and transparency. Implementing such safeguards requires integrating these principles into the training paradigm or embedding the model within frameworks~\cite{li2024satori}, which enforce comprehensive ethical checks. By positioning User-VLM 360° as a foundational component within such systems, it becomes possible to balance personalization with accountability, fostering ethically sound AI applications while maintaining adaptability for diverse user needs.

\section{Discussion and Future Work}
While User-VLM 360° has the potential to significantly enhance user experiences in healthcare, education, and assistive robotics, it also raises advanced technical considerations for future works.

\paragraph{Interactions with Multiple Parties} One limitation to discuss is that this work is primarily focused on dyadic interactions, involving a single robot and a single human. However, many social interactions involve multiple agents, such as a couple or a group of individuals. In scenarios where two people, such as a couple, are asking for a recommendation, the robot would need to consider the preferences and contexts of both individuals simultaneously. This introduces additional complexity, as the robot must balance and integrate the needs and preferences of multiple users to provide a coherent and satisfactory response.

\paragraph{Cognitive Metrics} Another important discussion point is the evaluation of human-robot interactions based on the subjective perception of the human user. While the User-VLM 360° framework demonstrates strong performance on objective benchmarks, such as F1-score, the human user's subjective experience is equally crucial. Factors like affiliation, trust, intimacy, and rapport play significant roles in determining the success and acceptance of human-robot interactions. Although these higher-level concepts are beyond the scope of this work, they are worth mentioning as they highlight the multifaceted nature of human-robot interactions and the need for future research to address these subjective aspects comprehensively.
\section{Conclusion}
Personalizing interactions between humans and robots equipped with vision-language models is essential for scalable and socially intelligent collaboration. Current methods often overlook individual nuances and raise ethical concerns due to biases in user data. To address this, we introduced User-VLM 360°, a framework that combines multimodal user context modeling with bias-aware optimization. This approach includes real-time adaptive tuning using visual, linguistic, and behavioral signals, bias mitigation, and a curated socio-emotive interaction dataset. Evaluations show significant improvements, and deployment on the Pepper robot confirms real-time adaptability.

\section{Impact Statement }
This paper introduces the User-VLM 360° framework, designed to advance personalized human-robot interactions by integrating VLMs into robotic systems. The framework focuses on user-aware tuning and bias mitigation to ensure ethical and fair responses, addressing concerns about data privacy, user consent, and safety. While this technology has the potential to significantly enhance user experiences in healthcare, education, and assistive robotics, it also raises ethical considerations and societal impacts that must be responsibly managed. These concerns include privacy risks, bias, and discrimination (such as stereotyping, exclusion, and fairness issues). However, thanks to a verification framework, explained in \Cref{app:verif}, many of these issues can be mitigated.

\section*{Acknowledgments}

The authors sincerely acknowledge the financial support of the French National Research Agency (ANR) for the ANITA project (Grant No. ANR-22-CE38-0012-01). We also extend our gratitude to \href{https://generated.photos}{Generated Photos} for generously providing 10,000 generated face entries.

\nocite{langley00}

\bibliography{ref}

\begin{thebibliography}{73}
\providecommand{\natexlab}[1]{#1}
\providecommand{\url}[1]{\texttt{#1}}
\expandafter\ifx\csname urlstyle\endcsname\relax
  \providecommand{\doi}[1]{doi: #1}\else
  \providecommand{\doi}{doi: \begingroup \urlstyle{rm}\Url}\fi

\bibitem[Agrawal et~al.(2024)Agrawal, Antoniak, Hanna, Bout, Chaplot, Chudnovsky, Costa, De~Monicault, Garg, Gervet, et~al.]{agrawal2024pixtral}
Agrawal, P., Antoniak, S., Hanna, E.~B., Bout, B., Chaplot, D., Chudnovsky, J., Costa, D., De~Monicault, B., Garg, S., Gervet, T., et~al.
\newblock Pixtral 12b.
\newblock \emph{arXiv preprint arXiv:2410.07073}, 2024.

\bibitem[Agrigoroaie \& Tapus(2016)Agrigoroaie and Tapus]{agrigoroaie2016developing}
Agrigoroaie, R.~M. and Tapus, A.
\newblock Developing a healthcare robot with personalized behaviors and social skills for the elderly.
\newblock In \emph{2016 11th ACM/IEEE International Conference on Human-Robot Interaction (HRI)}, pp.\  589--590. IEEE, 2016.

\bibitem[Alaluf et~al.(2025)Alaluf, Richardson, Tulyakov, Aberman, and Cohen-Or]{alaluf2025myvlm}
Alaluf, Y., Richardson, E., Tulyakov, S., Aberman, K., and Cohen-Or, D.
\newblock Myvlm: Personalizing vlms for user-specific queries.
\newblock In \emph{European Conference on Computer Vision}, pp.\  73--91. Springer, 2025.

\bibitem[Allam(2024)]{allam2024biasdpo}
Allam, A.
\newblock Biasdpo: Mitigating bias in language models through direct preference optimization.
\newblock \emph{arXiv preprint arXiv:2407.13928}, 2024.

\bibitem[An et~al.(2024)An, Yang, Lu, Zeng, Luo, Chen, Cao, Liang, She, Zhang, et~al.]{an2024mc}
An, R., Yang, S., Lu, M., Zeng, K., Luo, Y., Chen, Y., Cao, J., Liang, H., She, Q., Zhang, S., et~al.
\newblock Mc-llava: Multi-concept personalized vision-language model.
\newblock \emph{arXiv preprint arXiv:2411.11706}, 2024.

\bibitem[Andriella et~al.(2020)Andriella, Torras, and Alenya]{andriella2020short}
Andriella, A., Torras, C., and Alenya, G.
\newblock Short-term human--robot interaction adaptability in real-world environments.
\newblock \emph{International Journal of Social Robotics}, 12\penalty0 (3):\penalty0 639--657, 2020.

\bibitem[Bonci et~al.(2023)Bonci, Gaudeni, Giannini, and Longhi]{bonci2023robot}
Bonci, A., Gaudeni, F., Giannini, M.~C., and Longhi, S.
\newblock Robot operating system 2 (ros2)-based frameworks for increasing robot autonomy: A survey.
\newblock \emph{applied sciences}, 13\penalty0 (23):\penalty0 12796, 2023.

\bibitem[Cavallini et~al.(2021)Cavallini, Ceccato, Bertoglio, Francescani, Vigato, Ianes, and Lecce]{cavallini2021can}
Cavallini, E., Ceccato, I., Bertoglio, S., Francescani, A., Vigato, F., Ianes, A.~B., and Lecce, S.
\newblock Can theory of mind of healthy older adults living in a nursing home be improved? a randomized controlled trial.
\newblock \emph{Aging Clinical and Experimental Research}, 33:\penalty0 3029--3037, 2021.

\bibitem[Chen(2024)]{vlm_dpo_example}
Chen, A.
\newblock Vlm-dpo-example dataset, 2024.
\newblock URL \url{https://huggingface.co/datasets/alexchen4ai/vlm-dpo-example}.
\newblock Accessed: 2025-01-29.

\bibitem[Chen et~al.(2023)Chen, Li, Yan, Wang, Gunaratna, Yadav, Tang, Srinivasan, Zhou, Huang, et~al.]{chen2023AlpaGasus}
Chen, L., Li, S., Yan, J., Wang, H., Gunaratna, K., Yadav, V., Tang, Z., Srinivasan, V., Zhou, T., Huang, H., et~al.
\newblock Alpagasus: Training a better alpaca with fewer data.
\newblock \emph{arXiv preprint arXiv:2307.08701}, 2023.

\bibitem[Chen et~al.(2024{\natexlab{a}})Chen, Wu, Sun, Wang, Zhou, Cao, Ding, Zhao, Li, et~al.]{chen2024alignbot}
Chen, P., Wu, Z., Sun, J., Wang, D., Zhou, P., Cao, N., Ding, Y., Zhao, B., Li, X., et~al.
\newblock Alignbot: Aligning vlm-powered customized task planning with user reminders through fine-tuning for household robots.
\newblock \emph{arXiv preprint arXiv:2409.11905}, 2024{\natexlab{a}}.

\bibitem[Chen et~al.(2024{\natexlab{b}})Chen, Jie, and Ma]{chen2024llava}
Chen, S., Jie, Z., and Ma, L.
\newblock Llava-mole: Sparse mixture of lora experts for mitigating data conflicts in instruction finetuning mllms.
\newblock \emph{arXiv preprint arXiv:2401.16160}, 2024{\natexlab{b}}.

\bibitem[Deruyttere et~al.(2019)Deruyttere, Vandenhende, Grujicic, Van~Gool, and Moens]{deruyttere2019talk2car}
Deruyttere, T., Vandenhende, S., Grujicic, D., Van~Gool, L., and Moens, M.-F.
\newblock Talk2car: Taking control of your self-driving car.
\newblock \emph{arXiv preprint arXiv:1909.10838}, 2019.

\bibitem[Dong et~al.(2023)Dong, Zhang, Huang, Ji, Zhan, and Chen]{dong2023hubo}
Dong, Z., Zhang, W., Huang, X., Ji, H., Zhan, X., and Chen, J.
\newblock Hubo-vlm: Unified vision-language model designed for human robot interaction tasks.
\newblock \emph{arXiv preprint arXiv:2308.12537}, 2023.

\bibitem[Dubey et~al.(2024)Dubey, Jauhri, Pandey, Kadian, Al-Dahle, Letman, Mathur, Schelten, Yang, Fan, et~al.]{dubey2024llama}
Dubey, A., Jauhri, A., Pandey, A., Kadian, A., Al-Dahle, A., Letman, A., Mathur, A., Schelten, A., Yang, A., Fan, A., et~al.
\newblock The llama 3 herd of models.
\newblock \emph{arXiv preprint arXiv:2407.21783}, 2024.

\bibitem[Eapen \& Adhithyan(2023)Eapen and Adhithyan]{eapen2023personalization}
Eapen, J. and Adhithyan, V.
\newblock Personalization and customization of llm responses.
\newblock \emph{International Journal of Research Publication and Reviews}, 4\penalty0 (12):\penalty0 2617--2627, 2023.

\bibitem[Frith \& Frith(2005)Frith and Frith]{frith2005theory}
Frith, C. and Frith, U.
\newblock Theory of mind.
\newblock \emph{Current biology}, 15\penalty0 (17):\penalty0 R644--R645, 2005.

\bibitem[Gordon et~al.(2025)Gordon, Bitton, Shafir, Garg, Chen, Lischinski, Cohen-Or, and Szpektor]{gordon2025mismatch}
Gordon, B., Bitton, Y., Shafir, Y., Garg, R., Chen, X., Lischinski, D., Cohen-Or, D., and Szpektor, I.
\newblock Mismatch quest: Visual and textual feedback for image-text misalignment.
\newblock In \emph{European Conference on Computer Vision}, pp.\  310--328. Springer, 2025.

\bibitem[Goyal et~al.(2017)Goyal, Khot, Summers-Stay, Batra, and Parikh]{goyal2017making}
Goyal, Y., Khot, T., Summers-Stay, D., Batra, D., and Parikh, D.
\newblock Making the v in vqa matter: Elevating the role of image understanding in visual question answering.
\newblock In \emph{Proceedings of the IEEE conference on computer vision and pattern recognition}, pp.\  6904--6913, 2017.

\bibitem[Hu et~al.(2021)Hu, Shen, Wallis, Allen-Zhu, Li, Wang, Wang, and Chen]{hu2021lora}
Hu, E.~J., Shen, Y., Wallis, P., Allen-Zhu, Z., Li, Y., Wang, S., Wang, L., and Chen, W.
\newblock Lora: Low-rank adaptation of large language models.
\newblock \emph{arXiv preprint arXiv:2106.09685}, 2021.

\bibitem[Irawan(2024)]{vqa_nle_llava}
Irawan, P.~A.
\newblock Vqa-nle-llava dataset, 2024.
\newblock URL \url{https://huggingface.co/datasets/patrickamadeus/vqa-nle-llava}.
\newblock Accessed: 2025-01-29.

\bibitem[Irfan et~al.(2021)Irfan, Ramachandran, Spaulding, Kalkan, Parisi, and Gunes]{irfan2021lifelong}
Irfan, B., Ramachandran, A., Spaulding, S., Kalkan, S., Parisi, G.~I., and Gunes, H.
\newblock Lifelong learning and personalization in long-term human-robot interaction (leap-hri).
\newblock In \emph{Companion of the 2021 ACM/IEEE international conference on human-robot interaction}, pp.\  724--727, 2021.

\bibitem[Jafari \& Vassileva(2023)Jafari and Vassileva]{jafari2023ethical}
Jafari, E. and Vassileva, J.
\newblock Ethical issues in explanations of personalized recommender systems.
\newblock In \emph{Adjunct Proceedings of the 31st ACM Conference on User Modeling, Adaptation and Personalization}, pp.\  215--219, 2023.

\bibitem[Jevti{\'c} et~al.(2018)Jevti{\'c}, Valle, Aleny{\`a}, Chance, Caleb-Solly, Dogramadzi, and Torras]{jevtic2018personalized}
Jevti{\'c}, A., Valle, A.~F., Aleny{\`a}, G., Chance, G., Caleb-Solly, P., Dogramadzi, S., and Torras, C.
\newblock Personalized robot assistant for support in dressing.
\newblock \emph{IEEE transactions on cognitive and developmental systems}, 11\penalty0 (3):\penalty0 363--374, 2018.

\bibitem[Jiang et~al.(2023)Jiang, Sablayrolles, Mensch, Bamford, Chaplot, Casas, Bressand, Lengyel, Lample, Saulnier, et~al.]{jiang2023mistral}
Jiang, A.~Q., Sablayrolles, A., Mensch, A., Bamford, C., Chaplot, D.~S., Casas, D. d.~l., Bressand, F., Lengyel, G., Lample, G., Saulnier, L., et~al.
\newblock Mistral 7b.
\newblock \emph{arXiv preprint arXiv:2310.06825}, 2023.

\bibitem[Karkkainen \& Joo(2021)Karkkainen and Joo]{karkkainen2021fairface}
Karkkainen, K. and Joo, J.
\newblock Fairface: Face attribute dataset for balanced race, gender, and age for bias measurement and mitigation.
\newblock In \emph{Proceedings of the IEEE/CVF winter conference on applications of computer vision}, pp.\  1548--1558, 2021.

\bibitem[Kristen \& Sodian(2014)Kristen and Sodian]{kristen2014theory}
Kristen, S. and Sodian, B.
\newblock Theory of mind (tom) in early education.
\newblock \emph{Contemporary perspectives on research in theory of mind in early childhood education}, pp.\  291--320, 2014.

\bibitem[Langley(2000)]{langley00}
Langley, P.
\newblock Crafting papers on machine learning.
\newblock In Langley, P. (ed.), \emph{Proceedings of the 17th International Conference on Machine Learning (ICML 2000)}, pp.\  1207--1216, Stanford, CA, 2000. Morgan Kaufmann.

\bibitem[Lauren{\c{c}}on et~al.(2024)Lauren{\c{c}}on, Tronchon, Cord, and Sanh]{laurenccon2024matters}
Lauren{\c{c}}on, H., Tronchon, L., Cord, M., and Sanh, V.
\newblock What matters when building vision-language models?
\newblock \emph{arXiv preprint arXiv:2405.02246}, 2024.

\bibitem[Lewis et~al.(2020)Lewis, Perez, Piktus, Petroni, Karpukhin, Goyal, K{\"u}ttler, Lewis, Yih, Rockt{\"a}schel, et~al.]{lewis2020retrieval}
Lewis, P., Perez, E., Piktus, A., Petroni, F., Karpukhin, V., Goyal, N., K{\"u}ttler, H., Lewis, M., Yih, W.-t., Rockt{\"a}schel, T., et~al.
\newblock Retrieval-augmented generation for knowledge-intensive nlp tasks.
\newblock \emph{Advances in Neural Information Processing Systems}, 33:\penalty0 9459--9474, 2020.

\bibitem[Li et~al.(2023)Li, Wang, Wang, Ge, Ge, and Shan]{li2023seed}
Li, B., Wang, R., Wang, G., Ge, Y., Ge, Y., and Shan, Y.
\newblock Seed-bench: Benchmarking multimodal llms with generative comprehension.
\newblock \emph{arXiv preprint arXiv:2307.16125}, 2023.

\bibitem[Li et~al.(2024{\natexlab{a}})Li, Wu, Chan, Turakhia, Quispe, Li, Welch, Silva, and Qian]{li2024satori}
Li, C., Wu, G., Chan, G. Y.-Y., Turakhia, D.~G., Quispe, S.~C., Li, D., Welch, L., Silva, C., and Qian, J.
\newblock Satori: Towards proactive ar assistant with belief-desire-intention user modeling.
\newblock \emph{arXiv preprint arXiv:2410.16668}, 2024{\natexlab{a}}.

\bibitem[Li et~al.(2024{\natexlab{b}})Li, Goyal, Semedo, and Kolter]{li2024inference}
Li, K.~Y., Goyal, S., Semedo, J.~D., and Kolter, J.~Z.
\newblock Inference optimal vlms need only one visual token but larger models.
\newblock \emph{arXiv preprint arXiv:2411.03312}, 2024{\natexlab{b}}.

\bibitem[Lin(2004)]{lin-2004-rouge}
Lin, C.-Y.
\newblock {ROUGE}: A package for automatic evaluation of summaries.
\newblock In \emph{Text Summarization Branches Out}, pp.\  74--81, Barcelona, Spain, July 2004. Association for Computational Linguistics.
\newblock URL \url{https://aclanthology.org/W04-1013/}.

\bibitem[Liu et~al.(2024{\natexlab{a}})Liu, Li, Li, and Lee]{liu2024improved}
Liu, H., Li, C., Li, Y., and Lee, Y.~J.
\newblock Improved baselines with visual instruction tuning.
\newblock In \emph{Proceedings of the IEEE/CVF Conference on Computer Vision and Pattern Recognition}, pp.\  26296--26306, 2024{\natexlab{a}}.

\bibitem[Liu et~al.(2024{\natexlab{b}})Liu, Li, Wu, and Lee]{liu2024visual}
Liu, H., Li, C., Wu, Q., and Lee, Y.~J.
\newblock Visual instruction tuning.
\newblock \emph{Advances in neural information processing systems}, 36, 2024{\natexlab{b}}.

\bibitem[Liu et~al.(2024{\natexlab{c}})Liu, Zhang, Gao, Wang, and Wang]{liu2024vision}
Liu, S., Zhang, J., Gao, R.~X., Wang, X.~V., and Wang, L.
\newblock Vision-language model-driven scene understanding and robotic object manipulation.
\newblock In \emph{2024 IEEE 20th International Conference on Automation Science and Engineering (CASE)}, pp.\  21--26. IEEE, 2024{\natexlab{c}}.

\bibitem[Ma et~al.(2023)Ma, Hong, Gul, Gandhi, Gao, and Krishna]{ma2023crepe}
Ma, Z., Hong, J., Gul, M.~O., Gandhi, M., Gao, I., and Krishna, R.
\newblock Crepe: Can vision-language foundation models reason compositionally?
\newblock In \emph{Proceedings of the IEEE/CVF Conference on Computer Vision and Pattern Recognition}, pp.\  10910--10921, 2023.

\bibitem[Magri et~al.(2024)Magri, Amirian, and Chetouani]{magri2024upgrading}
Magri, P., Amirian, J., and Chetouani, M.
\newblock Upgrading pepper robot s social interaction with advanced hardware and perception enhancements.
\newblock \emph{arXiv preprint arXiv:2409.01036}, 2024.

\bibitem[Mataric(2023)]{mataric2023robot}
Mataric, M.
\newblock A robot just for you: Multimodal personalized human-robot interaction and the future of work and care.
\newblock In \emph{Proceedings of the 25th International Conference on Multimodal Interaction}, pp.\  2--3, 2023.

\bibitem[Ning et~al.(2024)Ning, Liu, Wu, Wu, Berlowitz, Prakash, Green, O'Banion, and Xie]{ning2024user}
Ning, L., Liu, L., Wu, J., Wu, N., Berlowitz, D., Prakash, S., Green, B., O'Banion, S., and Xie, J.
\newblock User-llm: Efficient llm contextualization with user embeddings.
\newblock \emph{arXiv preprint arXiv:2402.13598}, 2024.

\bibitem[Nocentini et~al.(2019)Nocentini, Fiorini, Acerbi, Sorrentino, Mancioppi, and Cavallo]{nocentini2019survey}
Nocentini, O., Fiorini, L., Acerbi, G., Sorrentino, A., Mancioppi, G., and Cavallo, F.
\newblock A survey of behavioral models for social robots.
\newblock \emph{Robotics}, 8\penalty0 (3):\penalty0 54, 2019.

\bibitem[Oertel et~al.(2020)Oertel, Castellano, Chetouani, Nasir, Obaid, Pelachaud, and Peters]{oertel2020engagement}
Oertel, C., Castellano, G., Chetouani, M., Nasir, J., Obaid, M., Pelachaud, C., and Peters, C.
\newblock Engagement in human-agent interaction: An overview.
\newblock \emph{Frontiers in Robotics and AI}, 7:\penalty0 92, 2020.

\bibitem[Onoe et~al.(2025)Onoe, Rane, Berger, Bitton, Cho, Garg, Ku, Parekh, Pont-Tuset, Tanzer, et~al.]{onoe2025docci}
Onoe, Y., Rane, S., Berger, Z., Bitton, Y., Cho, J., Garg, R., Ku, A., Parekh, Z., Pont-Tuset, J., Tanzer, G., et~al.
\newblock Docci: Descriptions of connected and contrasting images.
\newblock In \emph{European Conference on Computer Vision}, pp.\  291--309. Springer, 2025.

\bibitem[Ouyang et~al.(2022)Ouyang, Wu, Jiang, Almeida, Wainwright, Mishkin, Zhang, Agarwal, Slama, Ray, et~al.]{ouyang2022training}
Ouyang, L., Wu, J., Jiang, X., Almeida, D., Wainwright, C., Mishkin, P., Zhang, C., Agarwal, S., Slama, K., Ray, A., et~al.
\newblock Training language models to follow instructions with human feedback.
\newblock \emph{Advances in neural information processing systems}, 35:\penalty0 27730--27744, 2022.

\bibitem[Pandey \& Gelin(2018)Pandey and Gelin]{8409927}
Pandey, A.~K. and Gelin, R.
\newblock A mass-produced sociable humanoid robot: Pepper: The first machine of its kind.
\newblock \emph{IEEE Robotics \& Automation Magazine}, 25\penalty0 (3):\penalty0 40--48, 2018.
\newblock \doi{10.1109/MRA.2018.2833157}.

\bibitem[Photos(2024)]{generatedphotos}
Photos, G.
\newblock Generated photos: Unique, worry-free model photos, 2024.
\newblock URL \url{https://generated.photos/}.
\newblock Accessed: 2025-01-29.

\bibitem[Radford et~al.(2021)Radford, Kim, Hallacy, Ramesh, Goh, Agarwal, Sastry, Askell, Mishkin, Clark, et~al.]{radford2021learning}
Radford, A., Kim, J.~W., Hallacy, C., Ramesh, A., Goh, G., Agarwal, S., Sastry, G., Askell, A., Mishkin, P., Clark, J., et~al.
\newblock Learning transferable visual models from natural language supervision.
\newblock In \emph{International conference on machine learning}, pp.\  8748--8763. PMLR, 2021.

\bibitem[Radford et~al.(2023)Radford, Kim, Xu, Brockman, McLeavey, and Sutskever]{radford2023robust}
Radford, A., Kim, J.~W., Xu, T., Brockman, G., McLeavey, C., and Sutskever, I.
\newblock Robust speech recognition via large-scale weak supervision.
\newblock In \emph{International conference on machine learning}, pp.\  28492--28518. PMLR, 2023.

\bibitem[Rafailov et~al.(2024)Rafailov, Sharma, Mitchell, Manning, Ermon, and Finn]{rafailov2024direct}
Rafailov, R., Sharma, A., Mitchell, E., Manning, C.~D., Ermon, S., and Finn, C.
\newblock Direct preference optimization: Your language model is secretly a reward model.
\newblock \emph{Advances in Neural Information Processing Systems}, 36, 2024.

\bibitem[Rahimi et~al.(2025)Rahimi, Abrini, Khoramshahi, and Chetouani]{rahimi2025user}
Rahimi, H., Abrini, M., Khoramshahi, M., and Chetouani, M.
\newblock User-vlm: Llm contextualization with multimodal pre-trained user models.
\newblock \emph{The 39th Annual AAAI Conference on Artificial Intelligence}, 2025.

\bibitem[Ramesh(2024)]{face_bench_five_task_sample}
Ramesh, G.~V.
\newblock Face_bench_five_task_sample, 2024.
\newblock URL \url{https://huggingface.co/datasets/gp06aug/Face_Bench_Five_Task_Sample}.
\newblock Accessed: 2025-01-29.

\bibitem[Robinson et~al.(2023)Robinson, Tidd, Campbell, Kuli{\'c}, and Corke]{robinson2023robotic}
Robinson, N., Tidd, B., Campbell, D., Kuli{\'c}, D., and Corke, P.
\newblock Robotic vision for human-robot interaction and collaboration: A survey and systematic review.
\newblock \emph{ACM Transactions on Human-Robot Interaction}, 12\penalty0 (1):\penalty0 1--66, 2023.

\bibitem[Romeo et~al.(2022)Romeo, McKenna, Robb, Rajendran, Nesset, Cangelosi, and Hastie]{romeo2022exploring}
Romeo, M., McKenna, P.~E., Robb, D.~A., Rajendran, G., Nesset, B., Cangelosi, A., and Hastie, H.
\newblock Exploring theory of mind for human-robot collaboration.
\newblock In \emph{2022 31st IEEE International Conference on Robot and Human Interactive Communication (RO-MAN)}, pp.\  461--468. IEEE, 2022.

\bibitem[Sahu et~al.(2024)Sahu, Raut, Samant, Gorijala, Lakshminarayanan, and Bhaskar]{sahu2024pop}
Sahu, P.~P., Raut, A., Samant, J.~S., Gorijala, M., Lakshminarayanan, V., and Bhaskar, P.
\newblock Pop-vqa-privacy preserving, on-device, personalized visual question answering.
\newblock In \emph{Proceedings of the IEEE/CVF Winter Conference on Applications of Computer Vision}, pp.\  8470--8479, 2024.

\bibitem[Shen et~al.(2018)Shen, Pang, Weiss, Schuster, Jaitly, Yang, Chen, Zhang, Wang, Skerrv-Ryan, et~al.]{shen2018natural}
Shen, J., Pang, R., Weiss, R.~J., Schuster, M., Jaitly, N., Yang, Z., Chen, Z., Zhang, Y., Wang, Y., Skerrv-Ryan, R., et~al.
\newblock Natural tts synthesis by conditioning wavenet on mel spectrogram predictions.
\newblock In \emph{2018 IEEE international conference on acoustics, speech and signal processing (ICASSP)}, pp.\  4779--4783. IEEE, 2018.

\bibitem[Song et~al.(2024)Song, Liang, Payandeh, Raj, Xiao, and Manocha]{song2024vlm}
Song, D., Liang, J., Payandeh, A., Raj, A.~H., Xiao, X., and Manocha, D.
\newblock Vlm-social-nav: Socially aware robot navigation through scoring using vision-language models.
\newblock \emph{IEEE Robotics and Automation Letters}, 2024.

\bibitem[Steiner et~al.(2024)Steiner, Pinto, Tschannen, Keysers, Wang, Bitton, Gritsenko, Minderer, Sherbondy, Long, et~al.]{steiner2024paligemma}
Steiner, A., Pinto, A.~S., Tschannen, M., Keysers, D., Wang, X., Bitton, Y., Gritsenko, A., Minderer, M., Sherbondy, A., Long, S., et~al.
\newblock Paligemma 2: A family of versatile vlms for transfer.
\newblock \emph{arXiv preprint arXiv:2412.03555}, 2024.

\bibitem[Tam(2023)]{alexa_qa}
Tam, Z.~R.
\newblock Alexa-qa dataset, 2023.
\newblock URL \url{https://huggingface.co/datasets/theblackcat102/alexa-qa}.
\newblock Accessed: 2025-01-29.

\bibitem[Tanevska et~al.(2020)Tanevska, Rea, Sandini, Ca{\~n}amero, and Sciutti]{tanevska2020socially}
Tanevska, A., Rea, F., Sandini, G., Ca{\~n}amero, L., and Sciutti, A.
\newblock A socially adaptable framework for human-robot interaction.
\newblock \emph{Frontiers in Robotics and AI}, 7:\penalty0 121, 2020.

\bibitem[Team et~al.(2024)Team, Riviere, Pathak, Sessa, Hardin, Bhupatiraju, Hussenot, Mesnard, Shahriari, Ram{\'e}, et~al.]{team2024gemma}
Team, G., Riviere, M., Pathak, S., Sessa, P.~G., Hardin, C., Bhupatiraju, S., Hussenot, L., Mesnard, T., Shahriari, B., Ram{\'e}, A., et~al.
\newblock Gemma 2: Improving open language models at a practical size.
\newblock \emph{arXiv preprint arXiv:2408.00118}, 2024.

\bibitem[team(2024)]{mistral_Nemo}
team, M.~A.
\newblock Mistral nemo, 2024.
\newblock URL \url{https://mistral.ai/news/mistral-nemo}.
\newblock Accessed: 2024-01-29.

\bibitem[Touvron et~al.(2023)Touvron, Lavril, Izacard, Martinet, Lachaux, Lacroix, Rozi{\`e}re, Goyal, Hambro, Azhar, et~al.]{touvron2023llama}
Touvron, H., Lavril, T., Izacard, G., Martinet, X., Lachaux, M.-A., Lacroix, T., Rozi{\`e}re, B., Goyal, N., Hambro, E., Azhar, F., et~al.
\newblock Llama: Open and efficient foundation language models.
\newblock \emph{arXiv preprint arXiv:2302.13971}, 2023.

\bibitem[Tu(2024)]{human-face-emotions-roboflow}
Tu, Y.~T.
\newblock Human face emotions dataset.
\newblock \url{https://huggingface.co/datasets/tukey/human_face_emotions_roboflow}, sep 2024.
\newblock Accessed: 2025-01-29.

\bibitem[Wang et~al.(2024)Wang, Zhang, Dong, Fang, and Feng]{wang2024vlm}
Wang, B., Zhang, J., Dong, S., Fang, I., and Feng, C.
\newblock Vlm see, robot do: Human demo video to robot action plan via vision language model.
\newblock \emph{arXiv preprint arXiv:2410.08792}, 2024.

\bibitem[Wu et~al.(2024)Wu, Huang, and Wei]{wu2024mixture}
Wu, X., Huang, S., and Wei, F.
\newblock Mixture of lora experts.
\newblock \emph{arXiv preprint arXiv:2404.13628}, 2024.

\bibitem[Yeh et~al.(2023)Yeh, Russell, Sivic, Heilbron, and Jenni]{yeh2023meta}
Yeh, C.-H., Russell, B., Sivic, J., Heilbron, F.~C., and Jenni, S.
\newblock Meta-personalizing vision-language models to find named instances in video.
\newblock In \emph{Proceedings of the IEEE/CVF Conference on Computer Vision and Pattern Recognition}, pp.\  19123--19132, 2023.

\bibitem[Zhai et~al.(2023)Zhai, Mustafa, Kolesnikov, and Beyer]{zhai2023sigmoid}
Zhai, X., Mustafa, B., Kolesnikov, A., and Beyer, L.
\newblock Sigmoid loss for language image pre-training.
\newblock In \emph{Proceedings of the IEEE/CVF International Conference on Computer Vision}, pp.\  11975--11986, 2023.

\bibitem[Zhang et~al.(2024{\natexlab{a}})Zhang, Huang, Jin, and Lu]{zhang2024vision}
Zhang, J., Huang, J., Jin, S., and Lu, S.
\newblock Vision-language models for vision tasks: A survey.
\newblock \emph{IEEE Transactions on Pattern Analysis and Machine Intelligence}, 2024{\natexlab{a}}.

\bibitem[Zhang et~al.(2024{\natexlab{b}})Zhang, Cheng, Lu, Zhuo, Wang, Cao, Guo, She, and Zhang]{zhang2024cls}
Zhang, Q., Cheng, A., Lu, M., Zhuo, Z., Wang, M., Cao, J., Guo, S., She, Q., and Zhang, S.
\newblock [cls] attention is all you need for training-free visual token pruning: Make vlm inference faster.
\newblock \emph{arXiv preprint arXiv:2412.01818}, 2024{\natexlab{b}}.

\bibitem[Zhang et~al.(2019)Zhang, Kishore, Wu, Weinberger, and Artzi]{zhang2019bertscore}
Zhang, T., Kishore, V., Wu, F., Weinberger, K.~Q., and Artzi, Y.
\newblock Bertscore: Evaluating text generation with bert.
\newblock \emph{arXiv preprint arXiv:1904.09675}, 2019.

\bibitem[Zhou et~al.(2022)Zhou, Yang, Loy, and Liu]{zhou2022learning}
Zhou, K., Yang, J., Loy, C.~C., and Liu, Z.
\newblock Learning to prompt for vision-language models.
\newblock \emph{International Journal of Computer Vision}, 130\penalty0 (9):\penalty0 2337--2348, 2022.

\bibitem[Zhuang et~al.(2024)Zhuang, Sun, Yu, Qiang, Wang, Zhang, and Dai]{zhuang2024hydra}
Zhuang, Y., Sun, H., Yu, Y., Qiang, R., Wang, Q., Zhang, C., and Dai, B.
\newblock Hydra: Model factorization framework for black-box llm personalization.
\newblock \emph{arXiv preprint arXiv:2406.02888}, 2024.

\end{thebibliography}
\bibliographystyle{icml2024}


\appendix

\section{Data Construction Details}
\label{app:data}

Here, we provide a detailed discussion of the datasets we have constructed in more details, including their sources, preprocessing steps, and the rationale behind their design choices.

\subsection{PT datasets}
\paragraph{GenUser}

It includes approximately 10K synthetic image-text pairs, featuring human faces alongside user profile information from diverse demographic backgrounds. The dataset is generated by \enquote{\textit{generated.photos}} platform to ensure privacy and avoid using real personal data. To promote fairness, 
the entries are intentionally designed to represent a broad range of demographic groups, capturing diversity across key characteristics such as age, gender, and ethnicity. 
Each entry is accompanied by a JSON file integrating over 10 visual attributes that support a wide range of information about user profiles.
These features, alongside the images, are processed using a VLM (\enquote{GPT-4o}) to generate a one-paragraph user profile, providing a concise yet detailed description based on the inferred demographic and emotional attributes. The 10K entries in the dataset are split into three parts: 1K for validation, 1K for testing, and 8K for training, ensuring a balanced distribution across the dataset for training and model evaluation.

\paragraph{FairUser}
It approximately consists of 100K real-world text-image pairs derived from the FairFace dataset~\cite{karkkainen2021fairface}. The dataset entries are carefully curated to ensure balance, diversity, and accurate labeling across race, gender, and age categories. Based on this dataset, we designed a user profile feature using the following template: \enquote{The person appears to be {race class} {gender class}, approximately {age class} years old}. This template facilitates a structured and interpretable representation of demographic attributes for profiling tasks. The 100K entries in the dataset are split into three parts: 10K for validation, 10K for testing, and 80K for training, ensuring a balanced distribution across the dataset for training and model evaluation.
\balance

\subsection{Instruction datasets}
\paragraph{AlpaGasus-VQA} AlpaGasus dataset is an unofficial general-purpose dataset containing 10K question-answer pairs released by gpt4life that have demonstrated effectiveness in fine-tuning LLMs. For each question in AlpaGasus, we used LLMs to assign scores to each category of age, gender, and race, and select seven images based on the question context and matched user profile characteristics, then refined the answers using GPT-4o to align them with the user profiles. The following is an example of assignment and the prompt used for generation of the personalized response.

\paragraph{Alexa-VQA} The Alexa QA dataset includes over 100,000 pairs of question-answer entries, covering a wide range of topics and contexts. For our study, we selected a random subset of 20,000 entries from this dataset. Each of these selected QA pairs was then assigned to a user image sourced from the FairFace dataset, which provides a diverse set of human faces with demographic labels. To ensure that the responses were relevant and tailored to individual users, we applied a personalized approach by modifying the response based on the user's characteristics and context, using a consistent set of instructions. 

\subsection{DPO datasets}

\paragraph{BiasVision DPO}
Bias-DPO contains 1.2K entries that focus on addressing sexist, racist, controversial, and inappropriate questions. For each entry in Bias-DPO, we assign ten user profiles with corresponding images selected based on semantic similarity between the user profiles and the questions. The images are curated to ensure diversity across age, gender, and ethnicity, thereby reducing the risk of overfitting to specific demographic groups. 

\begin{graybox}
    \textbf{Instruction:} \textit{Imagine you are answering questions of \{UserProfile\}. Provide personalized respond according to the demographic, socio-emotive profile of the user to the following question:\{Question\}}
\end{graybox}

\begin{center}
\tiny
\begin{lstlisting}[style=json]
{
  "question": "Give three tips for staying healthy.",
  "age": [
    {
      "category": "0-2",
      "score": 0,
      "reason": "Too young to understand health tips."
    },    
    {
      "category": "20-29",
      "score": 0.8,
      "reason": "More likely to be interested in health and well-being."
    },    
  ],
  "gender": [
    {
      "category": "Male",
      "score": 0.5,
      "reason": "Men may show varied interest in health."
    },
    {
      "category": "Female",
      "score": 0.6,
      "reason": "Women tend to show higher interest in health and well-being."
    }
  ],
  "race": [
    {
      "category": "east asian",
      "score": 0.6,
      "reason": "Generally health-conscious but varies across groups."
    },
    {
      "category": "indian",
      "score": 0.6,
      "reason": "Generally health-conscious but varies across groups."
    },
  ]
}
\end{lstlisting}
\end{center}

\end{document}